\documentclass[letterpaper, 10 pt, conference]{ieeeconf}  % Comment this line out if you need a4paper

\IEEEoverridecommandlockouts                              % This command is only needed if 
\usepackage[resetlabels]{multibib}
\usepackage{subfiles}
\usepackage{wrapfig}
\usepackage{csquotes}
\usepackage{gensymb}
\usepackage{pifont}
\usepackage{textcomp}
\usepackage[table]{xcolor}
\usepackage{todonotes}
\usepackage{blindtext}
\usepackage{soul}
\usepackage{graphicx}
\usepackage{mathtools}
\usepackage{multirow}
\usepackage{booktabs}
\usepackage{array}
\usepackage{tabularx}
\usepackage[hang,flushmargin,symbol]{footmisc}
\usepackage{amsmath} 
\usepackage{amssymb} 
\usepackage{amsfonts}
\usepackage{bm} 
\usepackage{siunitx}
\usepackage{cancel}
\usepackage{microtype}
\usepackage{xcolor}
\usepackage[breaklinks,colorlinks]{hyperref}
\usepackage{caption}
\usepackage{subcaption}
\usepackage{cite}
\usepackage[export]{adjustbox}
\usepackage{cuted}
\usepackage{pgfplots}
\pgfplotsset{width=\linewidth,compat=1.9}
\definecolor{color_red}{RGB}{228,26,28}
\definecolor{color_blue}{RGB}{55,126,184}
\definecolor{color_green}{RGB}{77,175,74}
\definecolor{color_purple}{RGB}{152,78,163}
\definecolor{color_orange}{RGB}{255,127,0}
\definecolor{color_brown}{RGB}{166,86,40}
\definecolor{color_pink}{RGB}{247,129,191}
\newcites{S}{References}
\newcommand*{\SUP}{}%
\usepackage[capitalize]{cleveref}
\crefname{section}{Sec.}{Secs.}
\Crefname{section}{Section}{Sections}
\Crefname{table}{Table}{Tables}
\crefname{table}{Tab.}{Tabs.}

\definecolor{mygreen}{HTML}{00A64F}
\definecolor{myred}{HTML}{ED1B23}

\newcommand{\secref}[1]{Sec.~\ref{#1}}
\renewcommand{\eqref}[1]{Eq.~(\ref{#1})}
\newcommand{\figref}[1]{Fig.~\ref{#1}}
\newcommand{\tabref}[1]{Tab.~\ref{#1}}

\newcommand{\rebuttal}[1]{\textcolor{black}{#1}} % \textcolor{red}{#1}}% 

\renewcommand{\baselinestretch}{0.99}

\newcommand{\net}{TOPICS}

\title{\LARGE \bf
Taxonomy-Aware Continual Semantic Segmentation in\\Hyperbolic Spaces for Open-World Perception
}

\author{Julia Hindel, Daniele Cattaneo, and Abhinav Valada% <-this % stops a space
\thanks{This work was partly funded by the Deutsche Forschungsgemeinschaft (DFG, German Research Foundation) – SFB 1597 – 499552394.}
\thanks{Department of Computer Science, University of Freiburg, Germany 
}}

\begin{document}

\maketitle
\thispagestyle{empty}
\pagestyle{empty}

%%%%%%%%%%%%%%%%%%%%%%%%%%%%%%%%%%%%%%%%%%%%%%%%%%%%%%%%%%%%%%%%%%%%%%%%%%%%%%%%
\begin{abstract}
Semantic segmentation models are typically trained on a fixed set of classes, limiting their applicability in open-world scenarios. Class-incremental semantic segmentation aims to update models with emerging new classes while preventing catastrophic forgetting of previously learned ones. However, existing methods impose strict rigidity on old classes, reducing their effectiveness in learning new incremental classes. In this work, we propose Taxonomy-Oriented Poincaré-regularized Incremental-Class Segmentation (\net) that learns feature embeddings in hyperbolic space following explicit taxonomy-tree structures. This supervision provides plasticity for old classes, updating ancestors based on new classes while integrating new classes at fitting positions. Additionally, we maintain implicit class relational constraints on the geometric basis of the Poincaré ball. This ensures that the latent space can continuously adapt to new constraints while maintaining a robust structure to combat catastrophic forgetting. We also establish eight realistic incremental learning protocols for autonomous driving scenarios, where novel classes can originate from known classes or the background. Extensive evaluations of \net~on the Cityscapes and Mapillary Vistas~2.0 benchmarks demonstrate that it achieves state-of-the-art performance. We make the code and trained models publicly available at \url{http://topics.cs.uni-freiburg.de}\footnote{\rebuttal{Code will be published upon acceptance.}
The authors thank Kshitij Sirohi for technical discussions.}.
\end{abstract}

% \begin{IEEEkeywords}
% Deep Learning for Visual Perception, Computer Vision for Automation
% \end{IEEEkeywords}

%%%%%%%%%%%%%%%%%%%%%%%%%%%%%%%%%%%%%%%%%%%%%%%%%%%%%%%%%%%%%%%%%%%%%%%%%%%%%%%%

\section{Introduction}
\label{sec:introduction}

Automated vehicles rely on scene semantics predicted from online sensor data~\cite{mohan2022perceiving} as well as HD maps~\cite{greve2023collaborative} for safe navigation. The dominant paradigm for scene understanding exploits semantic~\cite{gosala2023skyeye} or panoptic segmentation models~\cite{kappeler2023few} trained on a dataset with a fixed number of predetermined semantic categories. 
%Self-driving vehicles require comprehensive scene understanding capabilities to safely navigate in dynamic environments. Semantic segmentation predicts object categories for every pixel of camera image which enables safe navigation in correlation to annotated objects and terrains in a scene. Commonly, an image model is trained on a static dataset with a fixed set of pixel-wise labels for this task.
However, such vehicles operate in an open-world scenario where training data with new object classes appear over time. While one line of research focuses on detecting unknown objects~\cite{mohan2024panoptic}, Class-Incremental Learning (CIL) aims to update the model with new classes at periodic timesteps~\cite{zhou2023deep}. On one hand, training a new model from scratch every time new classes appear is not only computationally inefficient but also requires past and present data to be available. On the other hand, simply updating a trained model with new data will result in catastrophic forgetting of old knowledge as the model will be biased towards new classes~\cite{cermelli2023comformer}. Consequently, CIL methods aim to balance observing characteristics of new classes while preserving patterns of formerly learned classes as the model is evaluated on all seen classes~\cite{zhou2023deep}.

Class-Incremental Semantic Segmentation (CISS) incorporates the background shift as an additional challenge. This phenomenon occurs as pixels that belong to old classes are labeled as background in new data samples~\cite{cermelli2020mib}. Consequently, CISS methods need to address label inconsistencies, catastrophic forgetting, and generalization on new classes at the same time. State-of-the-art CISS methods restrain the forgetting of old knowledge with data replay, network expansion, or regularization. The latter focuses on constraining features of the new model to imitate those of the prior model with direct feature distillation~\cite{Douillard2020PLOPLW} or frozen old class weights~\cite{zhang22_microseg,cha2021_ssul}. We find that these restrictions significantly hinder the plasticity of the model as old class features cannot evolve. 

Furthermore, most methods are tailored to the highly curated PascalVOC dataset which deviates significantly from densely annotated automated driving scenarios. While two to three object categories appear per image in PascalVOC, driving datasets typically contain over twenty different object categories, and fewer pixels are assigned to the background class. Additionally, all CISS methods assume that new classes originate from the prior background. This scenario is unrealistic for automated driving as a change in requirements for navigation could also entail bifurcations of previously observed classes for better decision-making, e.g., a model is initially trained to uniformly segment humans but later this ability needs to be extended to distinguish different vulnerable road users.\looseness=-1 

\begin{figure}
    \centering
    \includegraphics[width=0.9\linewidth]{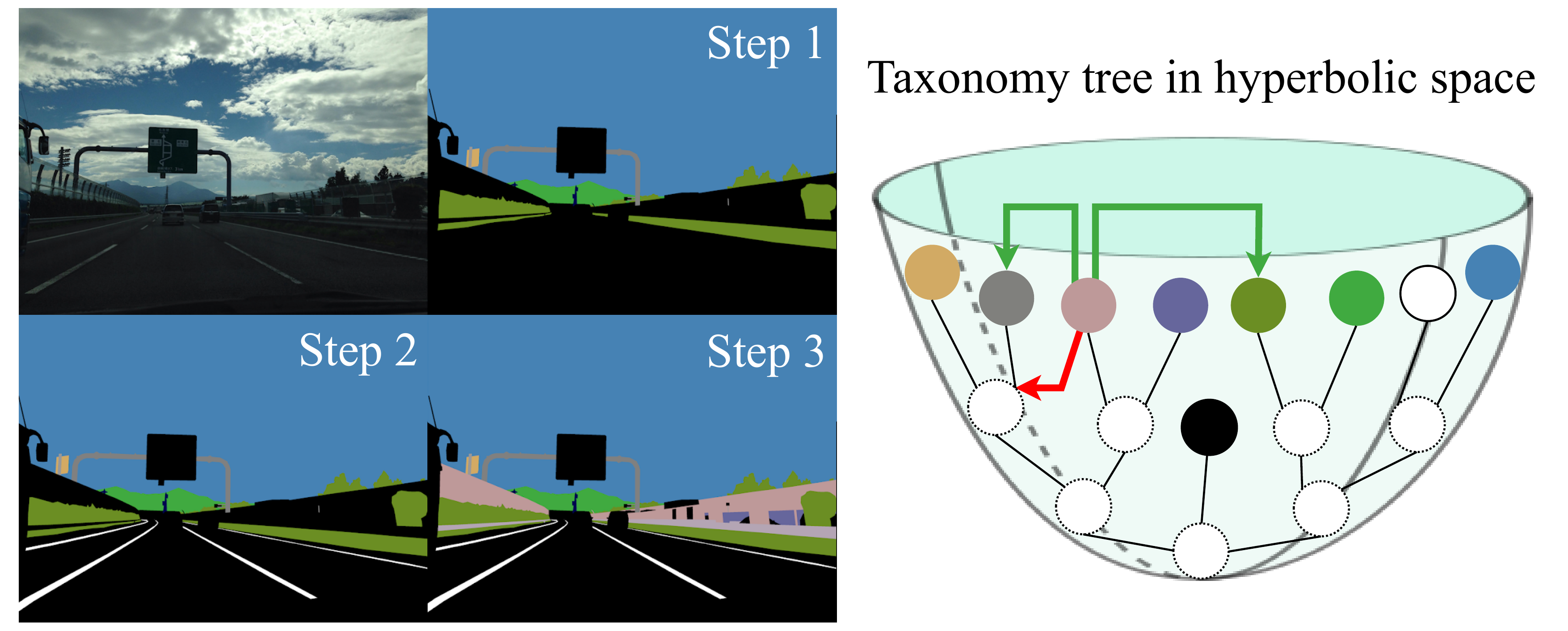}
    \caption{\net~leverages the explicit class taxonomy (black) and implicit relations (red and green) in hyperbolic space to balance rigidity and plasticity in taxonomic class-incremental semantic segmentation.}
    \label{fig:teaser}
    \vspace{-0.4cm}
\end{figure}

% Additionally, none of the observed methods take the class taxonomy of old and new classes into account when circumventing to forget old knowledge. We argue that incorporating the taxonomy of classes into the network model can act as an auxiliary regularization and eases learning new classes.

In this work, we introduce taxonomy-aware continual semantic segmentation for automated driving scenarios. Our proposed \textit{\textbf{T}axonomy-\textbf{O}riented \textbf{P}oincaré-regularized \textbf{I}ncremental \textbf{C}lass \textbf{S}egmentation} (\net) approach enforces features conform to taxonomy-tree structures in hyperbolic space. As a result, the overall class distribution is rigid and new classes are appended at fitting positions in \figref{fig:teaser}. \rebuttal{This supervision also allows old classes to evolve as the positions of ancestors are updated according to new classes which is a unique property of our method. We find that applying our distinct taxonomy-aware structure is beneficial for open-world scenarios since it effortlessly unites plasticity and rigidity.}
To further avoid catastrophic forgetting, we incorporate pseudo-labeling of the background and relation constraints of old class hyperplanes in \net. We argue that semantic classes inherit relations that go beyond the defined class taxonomy such as similar appearances or contexts. \rebuttal{We preserve consistency of these relations by maintaining relative mappings of prior class hyperplanes.} Lastly, we constrain features to have equidistant radii to maintain constant scarcity. We ensure that prior class features can only move in a circular direction around the hyperbolic center. Accordingly, new classes cannot result in a latent space shift of the complete taxonomy in their favor. We perform extensive evaluations of \net~on the standard Cityscapes~\cite{Cordts2016Cityscapes} and Mapillary Vistas~2.0~\cite{neuhold2017mapillary} benchmarks where it sets the new state-of-the-art. 

Our main contributions can thus be summarized as follows:
\begin{itemize}
    \item \net, a taxonomic-aware modeling in hyperbolic space to balance plasticity and rigidity for CISS.
    \item Two novel regularization losses tailored for incremental learning in hyperbolic space.
    \item Extensive evaluations and ablation study under eight different universal CISS settings for autonomous driving.
    \item Publicly available code and pretrained models at \url{http://topics.cs.uni-freiburg.de}.
\end{itemize}

\section{Related Work}
\label{sec:related-work}
In this section, we summarize existing works in class-incremental semantic segmentation, \rebuttal{hierarchical learning, and hyperbolic neural networks.}

{\parskip=3pt
\noindent\textit{Class-Incremental Semantic Segmentation}
% also regularization and 
\rebuttal{ (CISS) methods} rely on data replay, expansion, or distillation to avoid catastrophic forgetting. \rebuttal{In data replay methods, a subset of raw data~\cite{cha2021_ssul} or selected feature representations are stored in memory-buffers~\cite{chen2023STAR}. Further, replay data can also be recreated with GANs~\cite{maracani2021recall}.}
Expansion-based methods dedicate separate network components for particular semantic knowledge \rebuttal{in the form of parallel convolutions~\cite{zhang2022representation} or network branches~\cite{xiao2023endpoints}. The trainable branch is merged into the frozen branch after every incremental step to maintain consistent network sizes ~\cite{zhang2022representation, xiao2023endpoints}.}
% Another work first trains model on the novel categories before it fuses the prior model and temporary model in a third model. The authors claim to achieve better pseudo-labels by merging separately trained frozen and temporary model ~\cite{Yu2020SelfTrainingFC}.
Distillation approaches maintain prior model weights to restrain the current model for equivalent responses to the input data~\cite{zhou2023deep}. The pioneering approach MiB~\cite{cermelli2020mib} relates prior background logits to the combination of novel-class and background logits in the new model. This method is enhanced with gradient-based attribution weight initialization which identifies relevant classifier weights for novel classes from prior weights of the background class~\cite{goswami2023attribution}. On the other hand, PLOP~\cite{Douillard2020PLOPLW} labels the background with prior model predictions and distills pooled intermediate feature representations. Subsequent work focuses on learning an enhanced weighting term for distillation~\cite{wang24} or adapting this principle to transformer architectures~\cite{cermelli2023comformer, qiu2023sats}. The method SATS~\cite{qiu2023sats} also highlights the benefit of relation distillation between self-attention vectors in a SegFormer model. This weaker constraint allows the model to avoid forgetting while not constraining its plasticity. 
Prior work also trains segmentation models with sigmoid activation and binary cross-entropy loss as the instability of softmax activations hinders incremental learning~\cite{cha2021_ssul}. DKD~\cite{baek2022_dkd} further combines this approach with decoupled knowledge distillation while other approaches completely freeze the feature extractor and segment unknown background classes with saliency detectors~\cite{cha2021_ssul} or pre-trained models~\cite{zhang22_microseg}.}

However, all observed methods focus on novel classes from the background which significantly hinders their applicability in real-world scenarios where incremental learning could also entail a refinement of known classes. In this paper, we propose to simultaneously benchmark CISS methods for incremental learning from known classes and the background in the context of autonomous driving. Further, CISS methods do not \rebuttal{include any semantic relationships} between classes to balance plasticity and rigidity. We hypothesize that a hierarchical mapping of class features facilitates learning new classes while it constrains forgetting old classes. 

{\parskip=3pt
\noindent\textit{Hierarchical Learning} 
% Most segmentation problems only focus on training a flat classifier on leaf categories. 
\rebuttal{ methods outperform flat classifiers by modeling the semantic hierarchy in features~\cite{chen2023taxonomic} or combining logits with those of ancestor classes~\cite{liulei24, atigh2022hyperbolic}.
While hierarchical semantic segmentation primarily addresses \rebuttal{closed-set cases}, 
prior work in image classification concentrates on taxonomic class-incremental learning with network expansion~\cite{chen2023taxonomic} and replay-data~\cite{lin2022continual}. Lin~\textit{et~al.} first focused on taxonomic incremental semantic segmentation~\cite{lin2022continual}. However, they require all ancestor classes to be present in base training and allow using all history data which contradicts CISS principles. We introduce a more realistic form of taxonomic class-incremental semantic segmentation which allows increments from both background and known classes and complies with other CISS task definitions.}}

{\parskip=3pt
\noindent\textit{Hyperbolic Neural Networks}  \rebuttal{optimally} capture tree-like structures in text and graphs~\cite{ganea2018hnn}. For CNNs, the hyperbolic classification is modeled as a prototype-based approach~\cite{lang2022hyperbolic, cui2023_hypcil} or multinomial logistic regression~\cite{ganea2018hnn}. For the first, a cone entailment loss enforces all descendant prototypes to lie in the same geometric cone~\cite{desai2023meru}. In multinomial logistic regression, semantic classes are geometrically interpreted as hyperplanes, and hierarchies are explicitly modeled with a hierarchical softmax~\cite{atigh2022hyperbolic} or cosine margins~\cite{xu2023hier}. While hyperbolic neural networks have been extensively explored in image classification~\cite{xu2023hier, Spengler2023PoincarR} and metric learning~\cite{desai2023meru}, few prior works focus on semantic segmentation. Atigh~\textit{et~al.}~\cite{atigh2022hyperbolic} first shows the potential \rebuttal{of} hyperbolic multinomial regression for image segmentation. Follow-up work highlights the dense calibration capabilities of this network~\cite{weber2024flattening}.
% and leverages it for active learning~\cite{franco2024hyperbolic}. 
One pioneering work \rebuttal{~\cite{cui2023_hypcil} incorporates hyperbolic metric learning to enhance Euclidean-based knowledge retention for incremental image classification, which highlights the need for unique knowledge preservation methods that fully utilize the hyperbolic space}. Motivated by them, our work is \rebuttal{also} the first to explore hyperbolic spaces for CISS.

% image segmentation~\cite{chen2023unc} and 
% Further work extend euclidean models to operate in hyperbolic spaces such as convolutions~\cite{bdeir2024fully}, residual architectures~\cite{Spengler2023PoincarR} or Transformer architectures~\cite{ermolov2022hyperbolic}.

%  Modeling the semantic relationship of classes for semantic segmentation in the last layer has shown success where auxiliary ancestor classes are predicted~\cite{liulei24}.

% \cite{chen2023taxonomic}: explored taxonomic class incremental learning for image classification based on network expansion (course-to-fine learning). We maintain one feature extractor and balance stability and plasticity.

% \cite{lee2023online}: hierarchical label extension for image classification, based on rehearsal and memory management.

% \cite{lin2022continual}: Leco trains on evolving class ontologies where classes are always refined from old categories, keeps all history data. Semi-supervised learning with partial labels. 

% \cite{Abdelsalam2020IIRCII}: two hierarchy levels (coarse and fine) called IIRC (Incremental Implicitly-Refined Classification), model gets either or label (coarse or fine) for image classification. Hierarchy is implicit, methods rely on learning the hierarchy. Extend this work in learning separate latent spaces for each hierarchy level~\cite{zhao23}.
\section{Technical Approach} \label{sec:technical-approach}

\begin{figure*}
    \centering
    \includegraphics[width=0.8\linewidth]{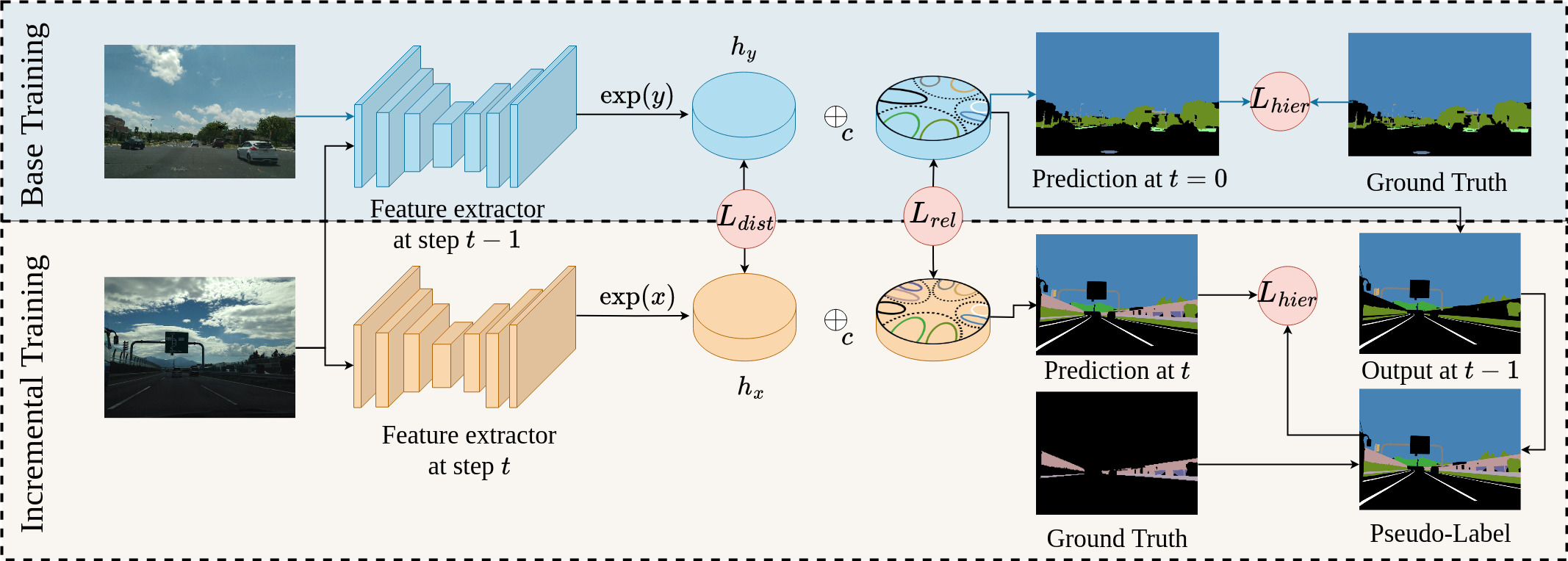}
    \caption{During base training of \net, features are mapped onto the Poincaré ball before the class hierarchy is explicitly enforced with $\mathcal{L}_{hier}$. In incremental steps, the old model is used to generate pseudo-labels of old classes \rebuttal{(PL)} and to regularize the last layer's weights with $\mathcal{L}_{rel}$ and feature radii with $\mathcal{L}_{dist}$.\looseness=-1}
    \label{fig:model}
    \vspace{-0.4cm}
\end{figure*}

\begin{figure}
    \centering
    \includegraphics[width=\linewidth]{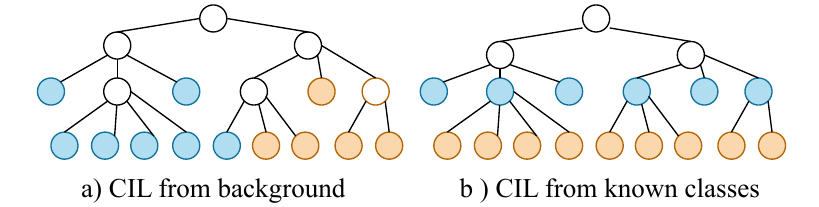}
    \caption{Visualization of the class taxonomic tree $\mathcal{H}$: a)~Novel classes originate from the background and b)~Novel classes originate from known classes. Base classes ($\mathcal{C}^{1}$) are colored in blue whereas novel classes ($\mathcal{C}^{2:T}$) are colored in orange. Novel ancestor nodes are visualized with orange outlines.}
    \label{fig:cil_def}
    \vspace{-0.4cm}
\end{figure}

In this section, we first introduce taxonomic CISS. We then present our \net\ approach which is tailored to incrementally learn new classes from either background or prior known classes as shown in \figref{fig:model}. We leverage the class taxonomy and implicit relations between prior classes to avoid catastrophic forgetting in incremental learning steps. We first train the model on the base dataset. The class hierarchy is explicitly enforced in the final network layer which is mapped in hyperbolic space. This geometric space ensures that classes are equidistant to each other irrespective of their hierarchy level which facilitates learning tree-like hierarchy structure. During the incremental steps, we leverage the old model's weights to create pseudo-labels for the background and employ the \rebuttal{class taxonomy, scarcity, and relation regularization losses to maintain explicit and implicit relations of old classes while learning the novel classes.}

\subsection{Class-Incremental Semantic Segmentation}
% ($\mathcal{C}^{1:t-1} \cap \mathcal{C}^{t} = \emptyset$)
CISS aims at training a model $f_\theta$ over $t = 1, ..., T$ incremental tasks. The first task ($t=1$) is denoted as base training while all subsequent tasks $t=2, ..., T$ are referred to as incremental steps. Every task is defined by its own disjoint label space $\mathcal{C}^t$ and training dataset $(x^t, y^t) \in \mathcal{D}^t$. $x^t$ refers to task-specific input images and $y^t \in \mathcal{Y}^t$  their pixelwise label according to $\mathcal{Y}^t = {b^t} \cup \mathcal{C}^t$. The background class $b^t$ includes all pixels whose true semantic class ($y$) is not included in $\mathcal{C}^t$. We consider the more realistic overlapped setting of CISS where training images ($x^t$) may include pixels whose dataset ground truth labels are old, current, or future classes. Their corresponding training label ($y^t$) is re-defined according to $\mathcal{Y}^t$.
% , i.e. $y^t = b^t \leftrightarrow y \in \mathcal{C}^{1:t-1} \cup \mathcal{C}^{t+1:T} \cup b$. 
After every task $t$, the network is challenged to make predictions on $\mathcal{C}^{1:t}$ whereas only true background pixels should not be associated with a semantic class.\looseness=-1

In comparison to~\cite{cha2021_ssul, Douillard2020PLOPLW, cermelli2020mib, zhang22_microseg, baek2022_dkd}, we do not constrain future classes to originate from the background in taxonomic CISS. We regard incremental scenarios where future classes are refinements of known classes or the background as shown in \figref{fig:cil_def}. We define $\mathcal{C}^{1:T}$ according to a class taxonomy tree $\mathcal{H}$ which has $l$ hierarchy levels and $N_L$ leaf nodes. 
Unlike taxonomic CIL as defined in~\cite{chen2023taxonomic}, we allow unbalanced trees and ancestor nodes to be introduced when its first leaf node $N_{L_{i}}$ is observed as illustrated in~\figref{fig:cil_def}a. Consequently, we generate more realistic incremental learning scenarios which are not restricted to breadth-first increments. When classes are refined from known classes (\figref{fig:cil_def}b), we define disjoint subsets $\mathcal{D}^t$ according to a fixed ratio, i.e. $\mathcal{Y}^t$ of every $x_i$ is constant and the same image cannot be observed with different labeling taxonomies at different time steps.

\subsection{Semantic Segmentation with the Poincaré Model}\label{sec:poincare}

\begin{figure}
    \centering
    \includegraphics[width=0.4\linewidth]{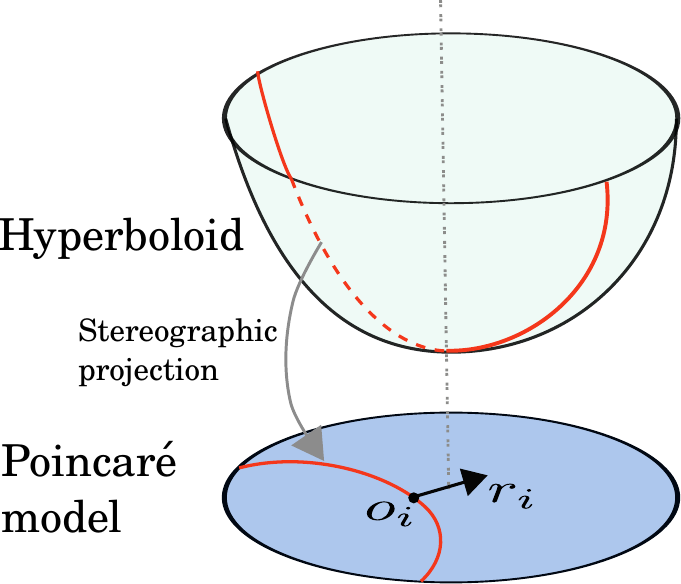}
    \caption{Visualization of a hyperplane on an upper sheet of a two-sheeted hyperboloid which is projected on a 2D Poincaré ball. The hyperplane has an offset of $o_i$ from the center and an orientation $r_i$.}
    \label{fig:poincare}
    \vspace{-0.4cm}
\end{figure}

We model the class hierarchy in hyperbolic space due to its favorable property of equidistant node connections on all hierarchy levels. Consequently, distances are inversely proportional to the semantic similarity of classes.
The hyperbolic space follows the geometry of constant negative curvature which is defined in the variable $c$. The Poincaré model is a stereographic projection of the upper sheet of a two-sheeted hyperboloid and is, therefore, represented by a unit ball as shown in~\figref{fig:poincare}. This hyperbolic model is formally defined by the manifold $\mathbb{D}_c^n=\left\{x \in \mathbb{R}^n: c\|x\|<1\right\}$ and Riemannian metric
\begin{equation}
g^{\mathbb{D}_c^n}=\left(\lambda_x^c\right)^2 g^{\mathbb{E}}=\left(\frac{2}{1-c\|x\|^2}\right)^2 g^{\mathbb{E}},
\end{equation}
where $g^{\mathbb{E}}$ is the euclidean tensor and $\lambda_x^c$ the conformal factor. When $c=0$, the Euclidean geometry is recovered.

For our \net~model, we only map the last neural network layer to hyperbolic space. Therefore, we first project the Euclidean features $e$ on the Poincaré ball at its origin which is defined as
\begin{equation}
h_i = \exp _0(e_i)=\tanh (\sqrt{c}\|e_i\|)(e_i /(\sqrt{c}\|e_i\|)).
\end{equation}

The geometric interpretation of multinomial regression in hyperbolic space suggests that every class $y$ is represented as a hyperplane in the Poincaré ball with offset $o_y \in \mathbb{D}_c^N$ and orientation $r_y \in {T}\mathbb{D}_c^N$(\figref{fig:poincare})
\begin{equation}
H_y^c=\left\{h_i \in \mathbb{D}_c^N,\left\langle-o_y \oplus_c h_i, r_y\right\rangle = 0 \right\}.
\end{equation}

Consequently, the likelihood of a class is defined as
\begin{equation}
p\left(\hat{y}=y \mid h_{i}\right) \propto \exp \left(\zeta_y\left(h_{i}\right)\right),
\end{equation}
where $\zeta_y$ is the signed distance of the feature $h_i$ to the hyperplane of $y$ which is approximated according to
{\footnotesize
\begin{equation} \label{eq:dist}
\zeta_y\left(h_{i}\right)=\frac{\lambda_{o_y}^c\left\|r_y\right\|}{\sqrt{c}} \sinh ^{-1}\left(\frac{2 \sqrt{c}\left\langle-o_y \oplus_c h_{i}, r_y\right\rangle}{\left(1-c\left\|-o_y \oplus_c h_{i}\right\|^2\right)\left\|r_y\right\|}\right)
\end{equation}}
with $\oplus_c$ being the Möbius addition
{\small
\begin{equation}
v \oplus_c w=\frac{\left(1+2 c\langle v, w\rangle+c\|w\|^2\right) v+\left(1-c\|v\|^2\right) w}{1+2 c\langle v, w\rangle+c^2\|v\|^2\|w\|^2}.
\end{equation}}

\subsection{Hierarchical Segmentation}
\label{sec:hierseg}

We model the hierarchy of semantic classes in the last layer of the network. We opt for a binary cross-entropy loss to ensure magnitudes of old and new class predictions do not correlate. Specifically, we extend the state-of-the-art hierarchical segmentation loss~\cite{liulei24} to multi-hierarchy levels and show its beneficial impact for incremental learning in hyperbolic space. Therefore, we model leaf nodes and all their ancestors as separate output classes $\mathcal{V}$ and use a combination of ancestor $\mathcal{A}$ and descendant $\mathcal{D}$ logits ($s$) in the loss function. We follow the tree-min loss~\cite{liulei24} defined as
\begin{align}
\mathcal{L}_{TM} = &\sum_{v \in \mathcal{V}}-l \log \left(\min _{u \in \mathcal{A}_v}\left(s_u\right)\right) \nonumber\\ 
&-\left(1-l\right) \log \left(1-\max _{u \in \mathcal{D}_v}\left(s_u\right)\right),
\end{align}
which penalizes hierarchical-inconsistent predictions for every class. For a correct prediction $l$, this loss penalizes the smallest logit of its ancestors $\mathcal{A}_v$. Conversely, for wrong predictions $1-l$, $\mathcal{L}^{\mathrm{TM}}$ punishes the maximum logits of its descendant $\mathcal{D}_v$ to reduce the score of complete root-to-leaf branches.
% If the ground-truth class $l$ equals class $a$ then 
% $\forall b \in \mathcal{A}_a\rightarrow P^{a} \leq P^{b}$. Similarly, for all classes $c$ who are unlike $l$, 
% $ \forall d \in \mathcal{P}_c\rightarrow 1 - P_{c} \leq 1 - P_{d}$ holds.
% commented as not used anymore
% Further, we apply a hierarchy-induced margin separation on the backbone features $\mathcal{L}^{\mathrm{M}}$. For every sample, positives are defined as classes on the same hierarchy level $l$ (i.e. siblings), whereas negatives are retrieved from any other hierarchy level. We note that with this formulation anchors and positive samples do not have aligned ancestors.

Further, we separately employ a categorical cross-entropy ($\mathcal{L}_{CE}$) on every hierarchy level. Therefore, we first retrieve 
$s_v' = \max _{u \in \mathcal{D}_v}\left(s_u\right)$ and convert the labels $y_t$ into unique binary labels for every hierarchy level $l$. This loss penalizes high prediction scores of sibling class descendants. The complete hierarchical loss is defined as
\begin{align}
\mathcal{L}_{hier} = \alpha \mathcal{L}_{TM} + \beta \mathcal{L}_{CE}.
% + \gamma \mathcal{L}^{\mathrm{CE}}.
\end{align}

% \begin{align}
% \mathcal{L}_{hier} = \alpha \mathcal{L}^{\mathrm{TM}} + \beta % \mathcal{L}^{\mathrm{M}} + \gamma \mathcal{L}^{\mathrm{CE}}.
% \end{align}

During inference, we multiply the logits of every leaf node $v_L$ with the logits of all its ancestors ($\mathcal{A}_{v_L}$) and remove non-leaf classes from the evaluation.

\subsection{Hierarchical Relation Distillation}
We reason that apart from explicit taxonomy relations between classes, an image model also captures implicit relations between classes in the form of relative similarity in feature space. Therefore, we aim to constrain these implicit relations to maintain relevant old class knowledge. In line with~\cite{Li_2024_WACV, qiu2023sats}, we argue that distilling relations is preferable to direct feature distillation as the latter restricts the model from re-distributing the feature space according to new classes. In comparison to the named prior work which distills relations based on feature maps, we propose to regularize the last layer's weights with this constraint. Consequently, we employ an InfoNCE loss on the hyperbolic class hyperplanes $\mathcal{H}_t$ to maintain closely grouped classes of the prior model in a similar constellation in the updated model. We define a distance between two classes $y^1$ and $y^2$ as the distance between one class offset $o_{y^1}$ and the hyperplane of the other class $H_{y^2}$.\looseness=-1

As outlined in \secref{sec:poincare}, an offset $o_{y^1}$ is a vector in hyperbolic space which defines the hyperplane of a class ${y^1}$ with orientation $r_{y^1}$. We retrieve absolute distances $d^{y^2}_{y^1}$ from signed distances which are computed using $\zeta_{y^2}(o_{y^1})$ as defined in \eqref{eq:dist}.
% We rescale the distances to positive values by applying the exponential function with base $e$. % Note that a large distance $d$ denotes a strong similarity of an offset to a hyperplane, whereas negative $d$ is re-scaled to small positive values.
Before beginning the training procedure, we utilize the old model's weights to compute the top $k$ most similar hyperplanes $H_y$ for every offset $o_y$ in $\mathcal{C}^{1:t-1}$. We neglect the background class in computing distances as we do not want to make constraints based on this variable class. 
% Further, we remove $\forall u \in \mathcal{P}$ from top $k$ as those classes are already constrained for similarity using the hierarchical loss as explained in \secref{sec:hierseg}. 
Further, we denote all positive anchors of a class, $k_{y_1}^{+}$, as the top $k$ smallest absolute distances to $o_{y_1}$ and enforce these relations to be maintained during the incremental training. We apply an InfoNCE-inspired loss:
\begin{align}
\mathcal{L}_{rel} = -\log \frac{\exp \left(1 - \tau  \cdot d_{k^{+}}/d_{max}\right)}{\sum_{i=0}^D \exp \left(1 - \tau  \cdot d_{i}/d_{max}\right).}
\end{align}
with $\tau$ being the temperature hyper-parameter.
With this constraint, we ensure relative implicit relations between old classes are maintained which creates an additional supervision for prior classes $\mathcal{C}^{1:t-1}$.

\subsection{Hyperbolic Distance Correlation}
Prior research highlights the correlation of the hyperbolic radius to the scarcity of observed features~\cite{franco2024hyperbolic} or the uncertainty of predictions in low dimensional space~\cite{atigh2022hyperbolic}. As incremental data is unbalanced with new classes appearing more frequently, we aim to constrain the radii of features to be unchanged between the old and new models. Therefore, we enforce features of the new and old model to be equidistant from the center of the Poincaré ball. We hypothesize that this constraint results in the hyperplanes of old classes rotating around the center and prevents a shift of the complete taxonomic tree in favor of the new classes. New space is allocated for new classes while the respective scarcity of old classes is not affected.

\begin{table*}
\centering
\caption{Continual semantic segmentation results on Cityscapes in mIoU (\%). Tasks defined as $\mathcal{C}^1$-$\mathcal{C}^T$($T$ tasks) and $h$ class hierarchy increments.}
\label{tab:city}
\begin{tabular}
{l|ccc|ccc|ccc|ccc}
 \toprule
 & \multicolumn{3}{c|}{\textbf{14-1 (6 tasks)}} & \multicolumn{3}{c|}{\textbf{10-1 (10 tasks)}} & \multicolumn{3}{c|}{\rebuttal{\textbf{7-5 (3 t.) -4 (2 t.)h}}} & \multicolumn{3}{c}{\textbf{7-18 (2 tasks)h}}\\
  \cmidrule{2-13}
\textbf{Method} & 1-14 & 15-19 & all & 1-10 & 11-19 & all & 1-7 & 8-25 & all & 1-7 & 8-25 & all\\
\midrule
PLOP~\cite{Douillard2020PLOPLW} & 63.54 & 15.38 & 48.33 & 60.75 & 27.97 & 42.96 & 88.56 & 18.14 & 20.75 & 88.73 & 15.06 & 17.99 \\
MiB~\cite{cermelli2020mib} & 66.37 & 14.36 & 50.05 & 61.80 & 32.97 & 45.73 & 77.66 & 6.61 & 9.83 & 90.10 & 5.71 & 9.64 \\
MiB + AWT~\cite{goswami2023attribution} & 65.60 & 19.19 & 50.72 & 60.97 & 35.70 & 46.55 & 84.65 & 10.46 & 13.64 & 90.19 & 5.61 & 9.56 \\
DKD~\cite{baek2022_dkd} & 68.83 & 14.70 & 51.86 & 66.77 & 34.52 & 48.92 & 89.46 & 0.56 & 4.98 & 89.19 & 4.29 & 8.32 \\
MicroSeg~\cite{zhang22_microseg} & 51.35 & 11.61 & 38.84 & 44.37 & 23.55 & 32.78 & 86.39 & 1.63 & 5.79 & 86.37 & 7.71 & 11.26 \\
\midrule
\net~(Ours) & \textbf{73.03} & \textbf{42.47} & \textbf{61.74} & \textbf{71.37} & \textbf{52.62} & \textbf{59.36} &
\textbf{90.02} & \textbf{51.31} & \textbf{50.69} & 
\textbf{90.33} & \textbf{61.62} & \textbf{59.98} \\
\hline
\end{tabular}
\vspace{-0.2cm}
\end{table*}

\begin{table*}
\centering
\caption{Continual semantic segmentation results on Mapillary Vistas~2.0 in mIoU (\%). Tasks defined as $\mathcal{C}^1$-$\mathcal{C}^T$($T$ tasks) and $h$ class hierarchy increments.}
\label{tab:map}
\setlength{\tabcolsep}{4pt}
\begin{tabular}{l|p{0.7cm}p{0.8cm}p{0.7cm}|p{0.7cm}p{0.8cm}p{0.7cm}|p{0.7cm}p{0.8cm}p{0.7cm}|p{0.7cm}p{0.8cm}p{0.7cm}}
 \toprule
 & \multicolumn{3}{c|}{\textbf{51-30 (3 tasks)}} & \multicolumn{3}{c|}{\textbf{71-10 (5 tasks)}} & \multicolumn{3}{c|}{\textbf{39-84 (2 tasks)h}} & \multicolumn{3}{c}{\textbf{39-21 (5 tasks)h}} \\
 \cmidrule{2-13}
\textbf{Method} & 1-51 & 52-111 & all & 1-71 & 72-111 & all & 1-39 & 40-123 & all & 1-39 & 40-123 & all \\
\midrule
PLOP~\cite{Douillard2020PLOPLW} & 20.83 & 8.97 & 14.59 & 18.12 & 5.74 & 13.83 & 19.15 & 5.52 & 9.51 & 17.79 & 3.14 & 6.64 \\
MiB~\cite{cermelli2020mib} & 16.72 & 11.48 & 13.77 & 15.10 & 8.43 & 12.58 & 19.38 & 13.61 & 11.64 & 16.49 & 6.51 & 8.86 \\
MiB + AWT~\cite{goswami2023attribution} & 18.33 & 13.27 & 15.89 & 15.78 & 9.69 & 13.91 & 19.76 & 13.47 & 15.47 & 17.75 & 11.16 & 12.84 \\
DKD~\cite{baek2022_dkd} & \textbf{25.49} & 12.82 & 18.74 & \textbf{22.71} & 11.03 & 18.65 & \textbf{24.24} & 3.20 & 9.15 & \textbf{28.04} & 1.53 & 8.06 \\
MicroSeg~\cite{zhang22_microseg} & 
9.39 & 4.19 & 6.62 & 8.38 & 3.46 & 6.65 & 12.70 & 1.07 & 4.41 & 13.69 & 0.80 & 4.08 \\ 
\midrule
\net~(Ours) & 23.76 &  \textbf{17.78} &  \textbf{20.35} &  22.10 & \textbf{14.55} & \textbf{19.20} & 24.16 & \textbf{21.49} & \textbf{21.94} & 26.67 & \textbf{22.57} & \textbf{23.35} \\
\hline
\end{tabular}
\vspace{-0.3cm}
\end{table*}

\section{Experimental Evaluation}

In this section, we present quantitative and qualitative
results of \net~on \rebuttal{eight} CISS settings in addition to a comprehensive ablation study to underline the importance of our contributions. Further, we detail the applied CISS settings and the training protocol that we employ.

\subsection{Datasets}
We evaluate \net~on the Cityscapes~\cite{Cordts2016Cityscapes} and Mapillary Vistas~2.0~\cite{neuhold2017mapillary} datasets. For both datasets, we define CISS protocols where incremental classes either primarily originate from the background or known classes. We only consider the more realistic case of overlapped CISS, where image pixels can belong to old, current, or future classes at any timestep.
The Cityscapes dataset consists of 19 semantic classes in addition to a void class. For CISS from the background, we adapt the 14-1 (6 tasks) and 10-1 (10 tasks) setting as proposed in~\cite{goswami2023attribution}. \rebuttal{The first 14 or 10 classes ($\mathcal{C}^1$) are learned during base training, while in this case, one class ($\mathcal{C}^T$) is added per incremental step. The task count ($T$) includes base training as the first task. For CISS from known classes, we learn 7 base classes that correspond to the official sub-categories defined for Cityscapes and increment the model in a 7-18 (2 tasks) or 7-5 (3 tasks) -4 (2 tasks) manner. }\looseness=-1

For Mapillary Vistas~2.0, we leverage 111 valid semantic classes and collapse all void pixels into one background class. Consequently, we define the settings of 51-30 (3 tasks), and 71-10 (5 tasks) for CISS from the background. On the other hand, we evaluate taxonomic incremental capabilities with 39-84 (2 tasks) and 39-21 (5 tasks) on this dataset.
For both datasets, we use the official validation split for testing and split the training data into training vs. validation with an 80:20 ratio. We note that the validation and test data remain constant for all incremental steps. For CISS from known classes, we divide the dataset into base and incremental dataset splits according to the number of learned classes within each step.
% add for mapillary increments are exclusive, for cityscapes they are not

\subsection{Experimental Setup} \label{sec:setup}
In line with prior work~\cite{Douillard2020PLOPLW, cermelli2020mib, zhang22_microseg}, we use the DeepLabV3 model with the ResNet-101 backbone which is pre-trained on ImageNet for all the experiments. We employ the Geoopt library~\cite{geoopt2020kochurov} to project the Euclidean features to a Poincaré ball with $c=2.0$ (different curvatures are explored in \secref{sec:curv}). Further, we follow the Möbius approximation defined in~\cite{atigh2022hyperbolic} for more efficient computations. We train \net~for 60 epochs per task with batch size $24$ for Cityscapes and $16$ for Mapillary Vistas 2.0 using the Riemannian SGD optimizer with momentum of $0.9$ and weight decay of $0.0001$. We use a poly learning rate scheduler with initial learning rates of $0.05$ for base training and $0.01$ in all incremental steps. We additionally ablate lower learning rates in \secref{sec:lr}. For the hierarchical loss function, we set the hyper-parameters to $\alpha=5$ and $\beta=1$ and ablate different hierarchical functions in \secref{sec:hh}. For Mapillary Vistas 2.0, we rescale the longest size to 2177 pixels before taking a non-empty crop of (1024,1024) and horizontal flipping. On the other hand, we train on random non-empty crops of (512,1024) with horizontal flipping for Cityscapes. Non-empty cropping biases image crops to include labeled masks (i.e., new classes) which could be neglected when applying random cropping.

\subsection{Quantitative Results}\label{sec:quan}
We compare \net~with five state-of-the-art CISS methods: PLOP~\cite{Douillard2020PLOPLW}, MiB~\cite{cermelli2020mib}, MiB+AWT~\cite{goswami2023attribution}, DKD~\cite{baek2022_dkd} and MicroSeg~\cite{zhang22_microseg}. For each method, we use the respective author's published code (\rebuttal{including hyperparameters}) and use the same augmentations outlined in \secref{sec:setup}. For Cityscapes, we train the PLOP~\cite{Douillard2020PLOPLW} on $512\times 512$ crops as the method is \rebuttal{limited} to squared input images. We evaluate the models using the mean intersection-over-union (mIoU) metric. Specifically, we evaluate the mIoU over all the base classes ($\mathcal{C}_1$) and novel classes ($\mathcal{C}_{2:T}$) separately as an indication of rigidity and plasticity. 
We present the results on Cityscapes in \tabref{tab:city}. On this dataset, \net~outperforms all baselines by at least $9.88$pp on the CISS from the background. While the difference in base IoU measures $4.2$pp, our method significantly exceeds the benchmarks by at least $16.9$pp in terms of novel IoU. This finding emphasizes that a balance between plasticity and rigidity is crucial to achieving superior results for class incremental learning. Further, we note the largest performance difference on incremental scenarios from known classes where \net~exceeds the best baseline by $29.94$pp in mIoU. \rebuttal{Thus}, we highlight the versatility of our method to balance plasticity and rigidity in all tested CISS settings. DKD~\cite{baek2022_dkd} achieves the highest benchmark on CISS from the background in our setting but does not generalize on the bifurcation of previously observed classes which we reason with frozen layers in incremental steps. MicroSeg~\cite{zhang22_microseg} also does not result in a favorable performance on Cityscapes and Mapillary Vistas 2.0 which is caused by low-quality proposals retrieved from the COCO dataset \rebuttal{and} fully freezing the backbone. \rebuttal{Further}, we present the mIoU over different increments for the 10-1 (10 tasks) scenario on Cityscapes in \figref{fig:miou10tasks}. We note that for most benchmarks, the mIoU increasingly deteriorates after increment \rebuttal{$T^3$}. \rebuttal{Conversely}, \net~achieves a constant performance from \rebuttal{$T^1$} to \rebuttal{$T^6$}, after which the performance again stabilizes. Therefore, our approach supremely maintains \rebuttal{prior class knowledge without restricting adaptability} to new classes.

\begin{figure} 
        \centering
        \resizebox{0.75\columnwidth}{!}{%
        \begin{tikzpicture} [font=\small]
        \begin{axis}[
            title={},
            ylabel={mIoU [\%]},
            legend style={font=\footnotesize},
            xlabel={Incremental step},
            xmin=0.5, xmax=10.5,
            ymin=28, ymax=93,
            ytick={30,40,50, 60, 70, 80},
            xtick={1,3,5,7,9,10},
            legend pos=north east,
            legend columns=3, 
            ymajorgrids=true,
            grid style=dashed,
            height=6cm
        ]

        \addplot[
            color=color_red,
            mark=square,
            line width=0.4mm,
            ]
            coordinates {
            (1,66.46)(2,68.28)(3,68.63)(4,67.38)(5,68.97)(6,67.24)(7,65.66)(8,60.24)(9,59.55)(10,59.36)
            };
        \addlegendentry{\net}
        \addplot[
            color=color_red,
            mark=x,
            line width=0.4mm,
            ]
            coordinates {
            (1,60.94)(2,63.56)(3,64.78)(4,63.41)(5,65.27)(6,64.21)(7,61.38)(8,56.71)(9,55.97)(10,55.83)
            };
        \addlegendentry{\net$_{b2}$}
        \addplot[
            color=color_brown,
            mark=triangle,
            line width=0.4mm,
            ]
            coordinates {
            (1,62.59)(2,64.98)(3,64.45)(4,59.35)(5,60.83)(6,58.26)(7,54.51)(8,49.93)(9,46.17)(10,45.73)
            };
        \addlegendentry{MiB\cite{cermelli2020mib}}
                \addplot[
            color=color_green,
            mark=star,
            line width=0.4mm,
            ]
            coordinates {
            (1,62.59)(2,64.95)(3,64.59)(4,60.11)(5,61.34)(6,58.23)(7,55.93)(8,50.23)(9,48.36)(10,46.55)
            };
        \addlegendentry{MiB+AWT~\cite{goswami2023attribution}}
        \addplot[
            color=color_purple,
            mark=o,
            line width=0.4mm,
            ]
            coordinates {
            (1,58.57)(2,61.44)(3,61.2)(4,56.63)(5,56.92)(6,55.31)(7,51.87)(8,48.01)(9,45.13)(10,42.96)
            };
        \addlegendentry{PLOP~\cite{Douillard2020PLOPLW}}
                 \addplot[
            color=color_orange,
            mark=diamond,
            line width=0.4mm,
            ]
            coordinates {
            (1,63.76)(2,65.55)(3,65.93)(4,61.30)(5,62.55)(6,60.32)(7,56.47)(8,51.84)(9,50.30)(10,48.92)
            };
        \addlegendentry{DKD~\cite{baek2022_dkd}}
        \addplot[
            color=color_blue,
            mark=x,
            line width=0.4mm,
            ]
            coordinates {
            (1,48.04)(2,46.27)(3,46.21)(4,41.77)(5,44.27)(6,41.94)(7,40.78)(8,39.06)(9,34.47)(10,32.78)
            };
        \addlegendentry{Microseg~\cite{zhang22_microseg}}
    \end{axis}
    \end{tikzpicture}}
\caption{Performance at every increment on Cityscapes 10-1 (10 task) setting. \rebuttal{\net$_{b2}$ is trained for half iterations in base training to illustrate knowledge retention with lower base training performance.}}
\label{fig:miou10tasks}
\vspace{-0.4cm}
\end{figure}

% results map
We present the results on Mapillary Vistas 2.0 in~\tabref{tab:map}. \net~outperforms all baselines by at least $1.6$pp on all CISS benchmarks. We justify these results with the rigidity of our model towards base classes while not restraining the learning of new knowledge. Our method significantly outperforms the baselines on novel IoU where we record an improvement of at least $3.5$pp. Further, the baselines significantly underperform on the CISS from known classes setting which shows the need for solutions tailored for both scenarios. We observe that \net~seemingly integrated the CISS from known classes into its hierarchical learning paradigm. We further discuss the performance on base classes in the supplementary material \ifdefined \SUP \secref{sec:quanap}\else Sec. \textcolor{red}{S.1}\fi.\looseness=-1

\subsection{Ablation Study}
In this section, we analyze the impact of various architectural components and hyperparameters on the performance of our approach. We perform all the ablation experiments on the Cityscapes 14-1 (6 tasks) setting.

\subsubsection{Influence of Different Components}
We show in~\tabref{tab:ab_elem} that \rebuttal{pseudo-labeling of the background prevents the collapse of base class performance when training with sigmoid activations and binary cross-entropy on flat labels. Further,} hyperbolic spaces are better calibrated as the performance after base training increases by $0.79$pp which complements findings in~\cite{weber2024flattening}. \rebuttal{The} performance improvement increases to $2.52$pp in mIoU after the final step which can be attributed to a difference of $5.73$pp in novel classes. Consequently, we suggest that equidistant class mappings in hyperbolic space support incremental learning.
Further, we find that our hierarchical loss results in a performance improvement of $1.85$pp which can be attributed to a performance increase in base and novel classes. Base classes benefit from the rigidity of the feature space as well as a high quality of learned features in base training ($1.34$pp in comparison to the flat classifier) which helps prevent forgetting. Additionally, the generalization on new classes is amplified by $6.52$pp. We reason that learning new classes is eased in a hierarchy as their mapping is already defined by trained ancestor nodes. \rebuttal{Thus}, we motivate hierarchical modeling in neural networks for open-world learning. Our regularization losses increase the performance by $0.21$pp and $0.14$pp, respectively. \rebuttal{Newly} added classes significantly benefit from our constant sparsity constraint ($0.73$pp) as their relative frequency largely differs in incremental steps. \rebuttal{Constraining} implicit class relations further improve the retention of new classes by $0.41$pp. \rebuttal{Further results are in the supplementary material \ifdefined \SUP \secref{sec:sup_reg}\else Sec. \textcolor{red}{S.5}\fi.}

\begin{table}
\centering
\caption{Ablation study on the efficacy of various components of \net~. All results are reported on Cityscapes in mIoU (\%). The setting $1-14_0$ represents the mIoU on base classes at time $t=0$. \rebuttal{PL: pseudo-labeling of the background with the old model.}}
\label{tab:ab_elem}
\begin{tabular}
{p{0.85cm}|p{0.25cm}p{0.4cm}p{0.4cm}p{0.45cm}|p{0.7cm}p{0.6cm}p{0.7cm}p{0.5cm}}
 \toprule
 & \multicolumn{8}{c}{\textbf{14-1 (6 tasks)}}\\
\textbf{Space} & \rebuttal{\textbf{$PL$}} & \textbf{$\mathcal{L}_{hier}$} & \textbf{$\mathcal{L}_{dist}$} & \textbf{$\mathcal{L}_{rel}$}   & 1-14$_{0}$ & 1-14 & 15-19 & all \\
\midrule
\multirow{2}{*}{\shortstack[l]{Eucl.}} & & & & & \rebuttal{72.72} & \rebuttal{0.00} & \rebuttal{14.55} & \rebuttal{3.64} \\
& \checkmark & & & & 72.72 & 71.06 & 29.09 & 57.02 \\
\midrule
\multirow{4}{*}{\shortstack[l]{Hyperb.}} & \checkmark & & & & 73.51 & 72.62 & 34.82 & 59.54 \\
& \checkmark & \checkmark & & & 74.85 & 72.94 & 41.34 & 61.39 \\
& \checkmark & \checkmark & \checkmark & & 74.85 & 72.98 & 42.06 & 61.60 \\
& \checkmark & \checkmark &\checkmark & \checkmark & \textbf{74.85} & \textbf{73.03} & \textbf{42.47} & \textbf{61.74}\\
% 73.04 & 39.74 & 61.06 
\hline
\end{tabular}
\vspace{-0.3cm}
\end{table}

\subsubsection{Learning Rate and Epochs} \label{sec:lr}
Some benchmarks adopt a lower learning rate~\cite{Douillard2020PLOPLW, cermelli2020mib, baek2022_dkd, goswami2023attribution} and train for less epochs~\cite{Douillard2020PLOPLW, goswami2023attribution, cermelli2020mib} during incremental steps \rebuttal{to prevent forgetting}. While our model benefits from the selected settings, we additionally show in \tabref{tab:lr} that \net~also outperforms all benchmarks with their selected hyperparameters. We note that training our model for shorter epochs or a lower learning rate significantly decreases the generalization on novel classes ($\mathcal{C}_{1:T}$). Further, the performance on base classes is affected which we reason with false positive segmentation of base classes. When training for $30$ epochs on a learning rate of $0.01$ this effect seems diminished as the base performance peaks with this reduced number of iterations.

\begin{table}
\centering
\caption{Influence of different learning rates and number of epochs during training increments on Cityscapes in mIoU (\%).}
\label{tab:lr}
\begin{tabular}
{ll|ccc}
 \toprule
 \multicolumn{5}{c}{\textbf{14-1 (6 tasks)}} \\
\textbf{Learning Rate} & \textbf{Epochs} &1-14 & 15-19 & all \\
\midrule
0.001 & 30 & 71.72 & 21.07 & 55.47\\
0.001 & 60 & 72.72 & 27.43 & 57.76\\
0.01 & 30 & \textbf{73.19} & 28.74 & 58.42 \\
0.01 &  60 & 73.03 & \textbf{42.47} & \textbf{61.74}\\
\hline
\end{tabular}
\vspace{-0.3cm}
\end{table}

\begin{figure*}
\centering
\begin{subfigure}[b]{0.495\textwidth}
\vskip 0pt
\centering
\footnotesize
\setlength{\tabcolsep}{0.05cm}% for the horiz padding
{
\renewcommand{\arraystretch}{0.2}% for the vertical padding
\newcolumntype{M}[1]{>{\centering\arraybackslash}m{#1}}
\begin{tabular}{cM{1.95cm}M{1.95cm}M{1.95cm}M{1.95cm}}
& Input image & DKD~\cite{baek2022_dkd} & \net~(Ours) & Improv./Error \\
(i) & \includegraphics[width=\linewidth, frame]{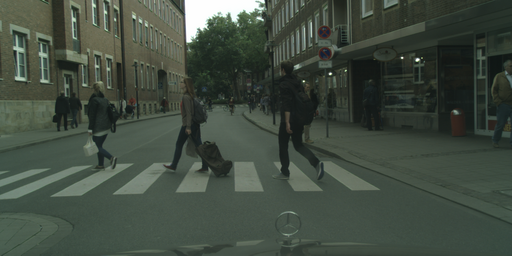} & \includegraphics[width=\linewidth, frame]{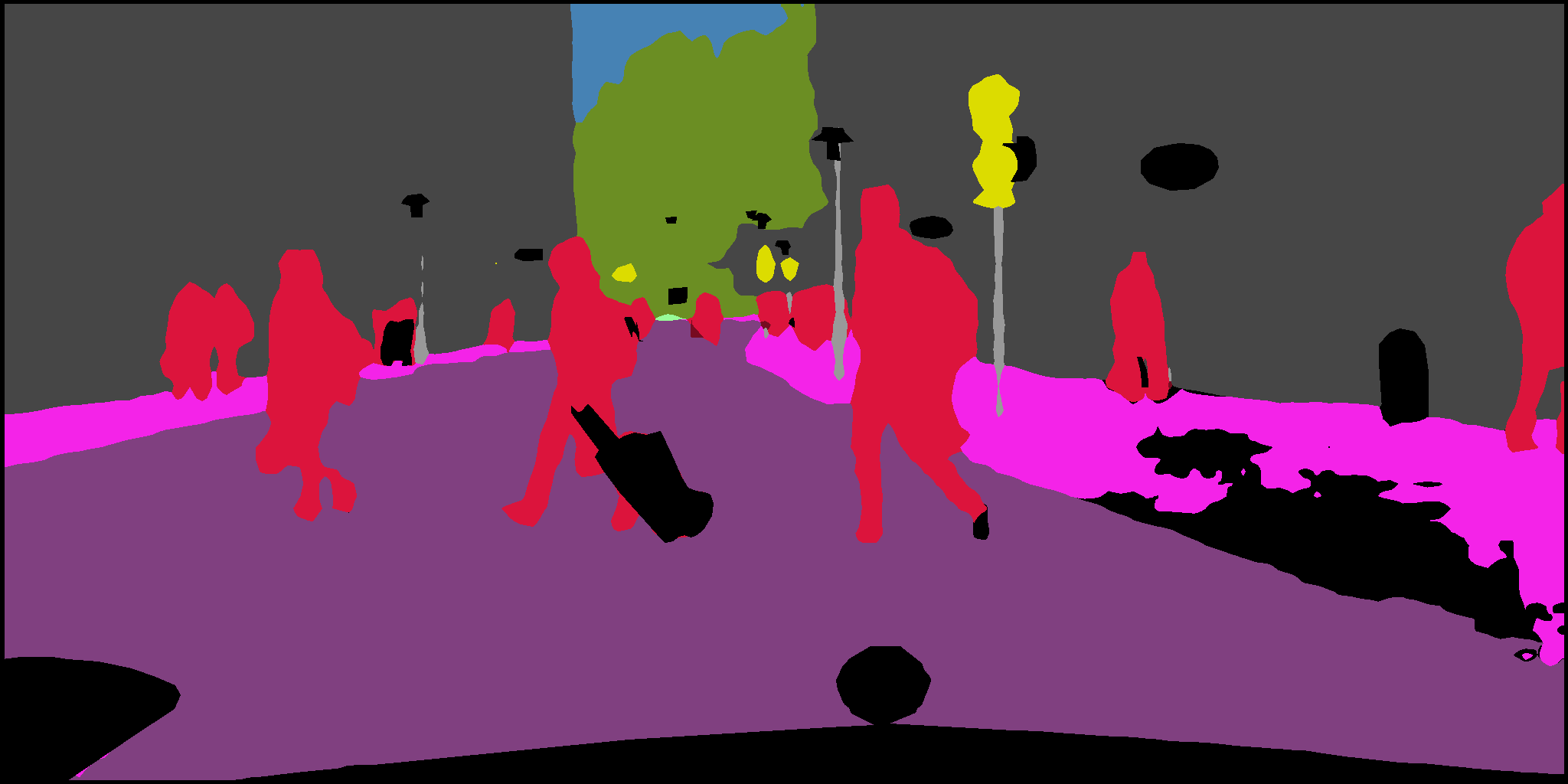} & \includegraphics[width=\linewidth, frame]{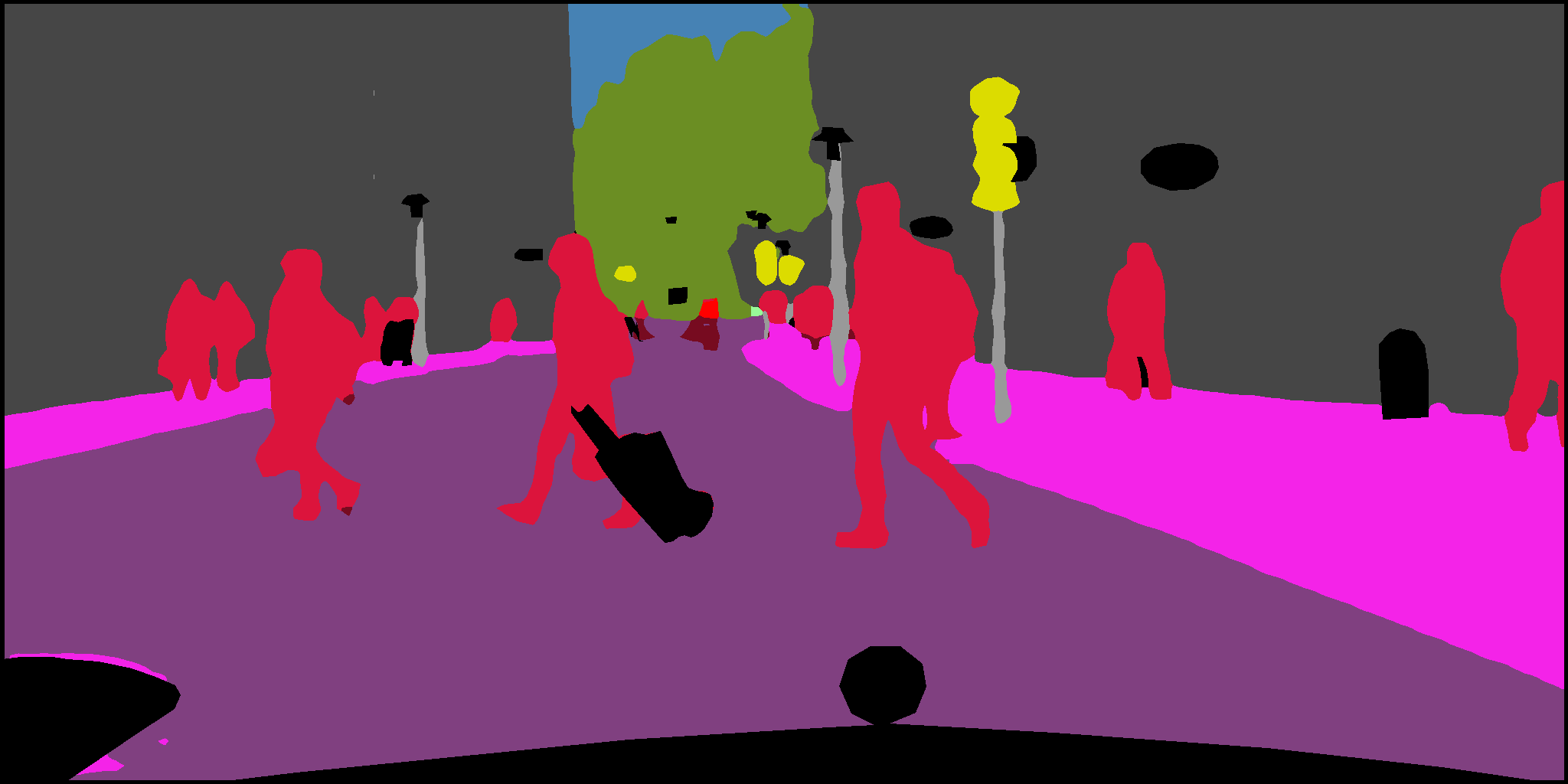} & \includegraphics[width=\linewidth, frame]{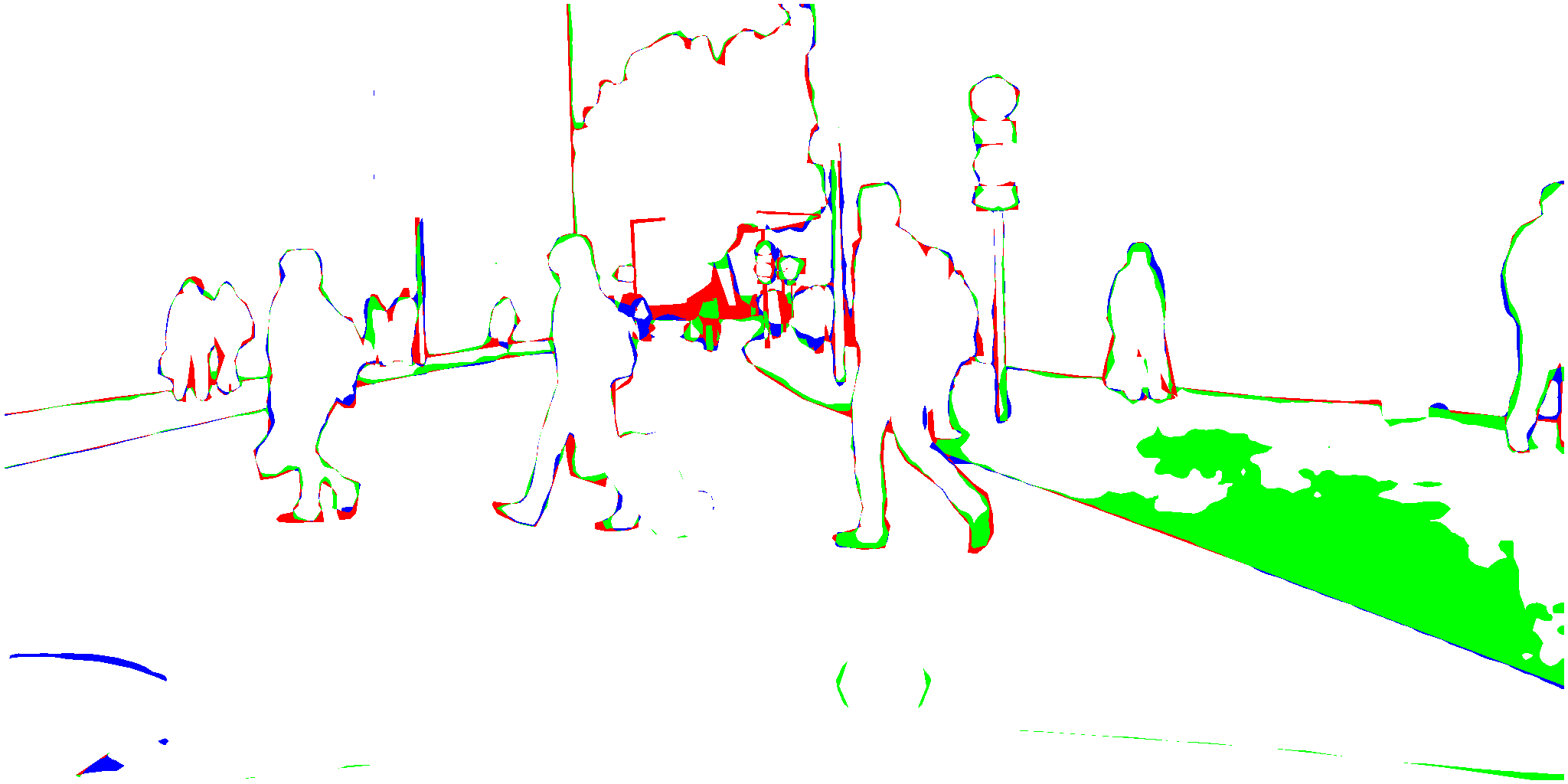}\\
\\
(ii) &  \includegraphics[width=\linewidth, frame]{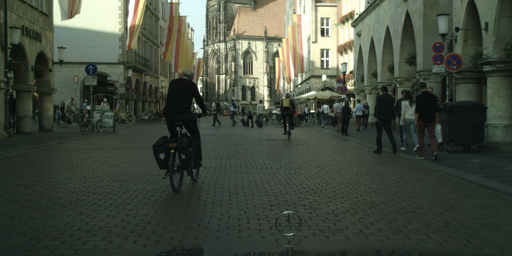} & \includegraphics[width=\linewidth, frame]{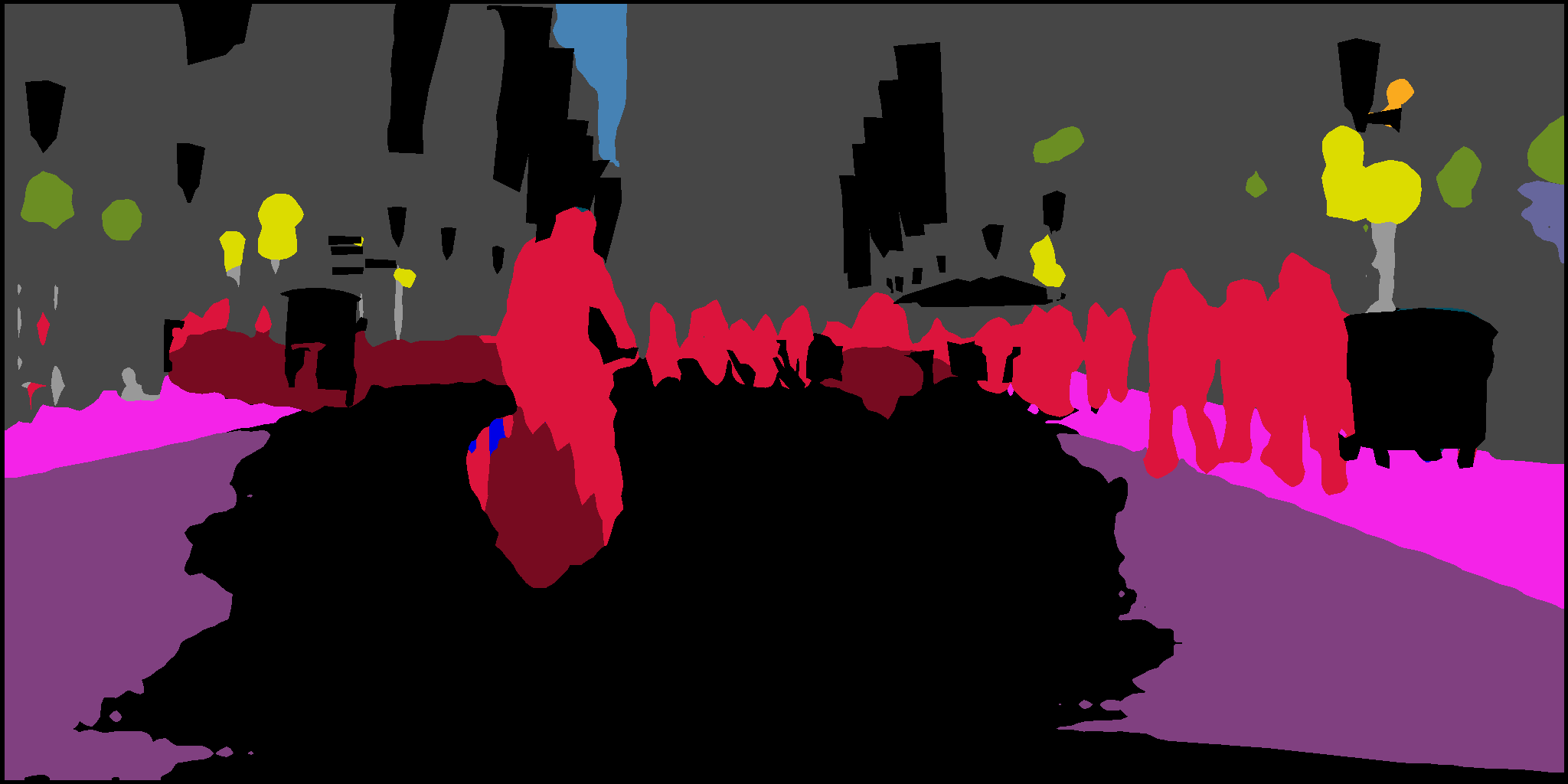} & \includegraphics[width=\linewidth, frame]{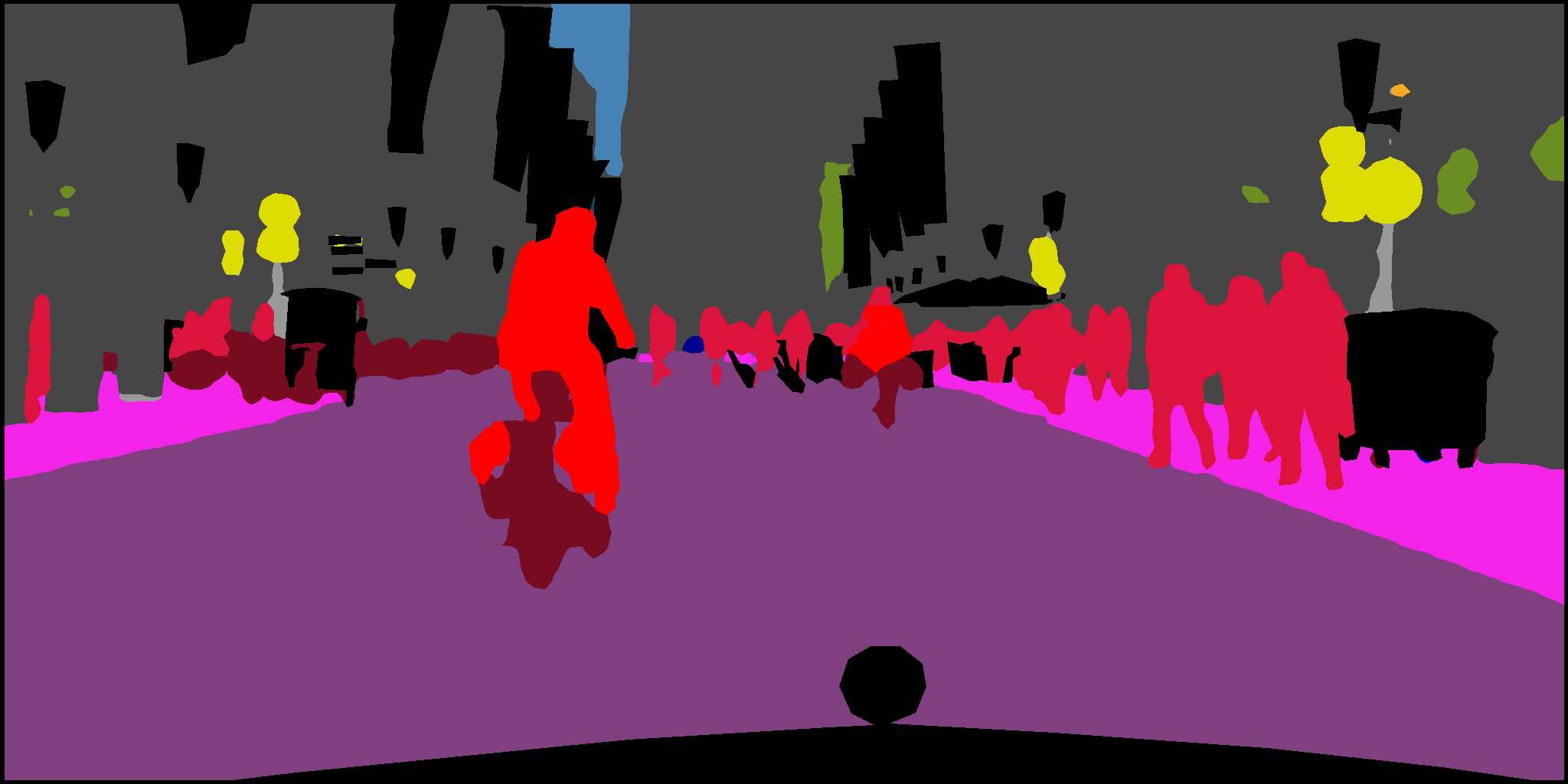} & \includegraphics[width=\linewidth, frame]{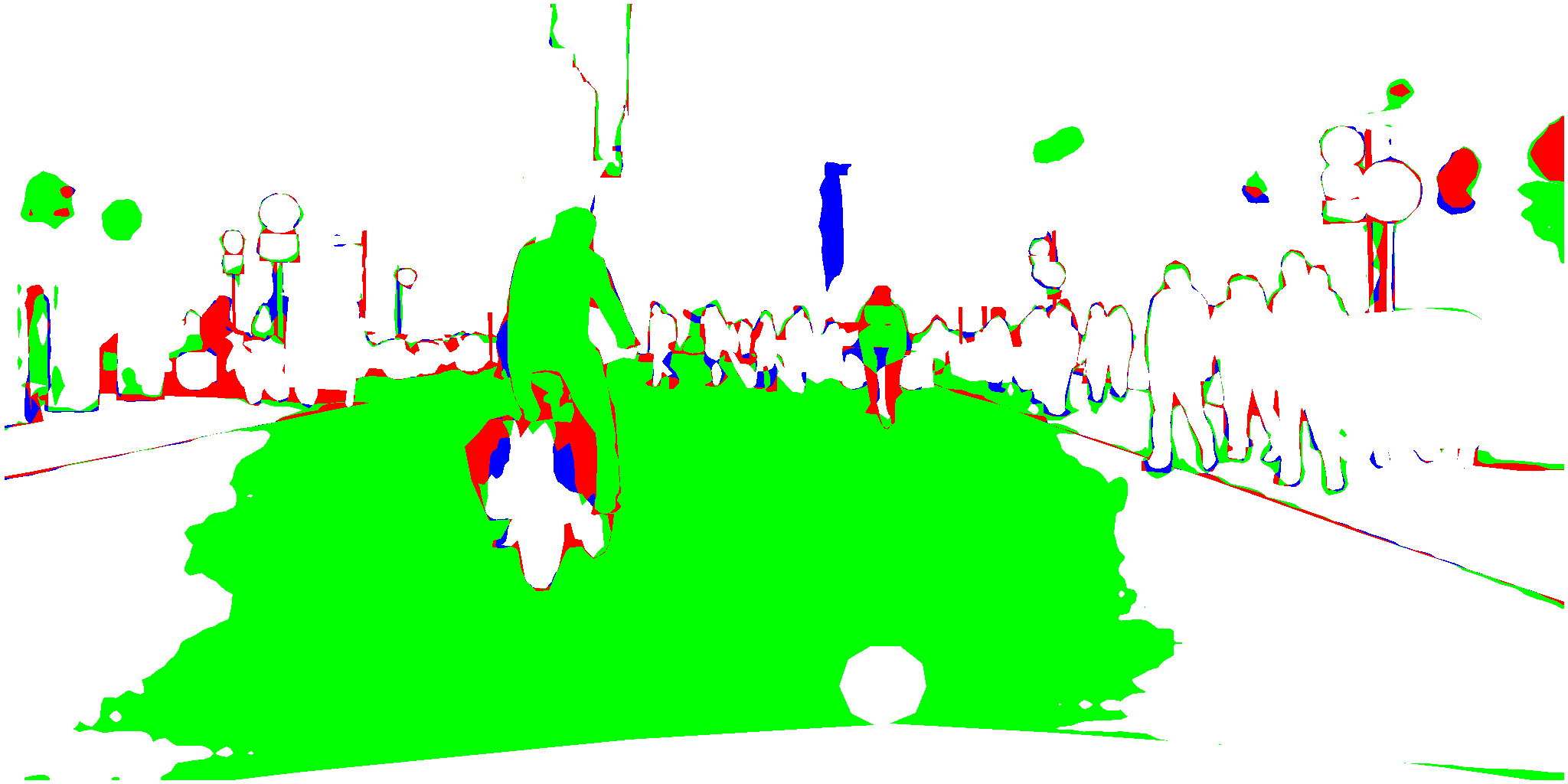}\\
\\
(iii) & \includegraphics[width=\linewidth, frame]{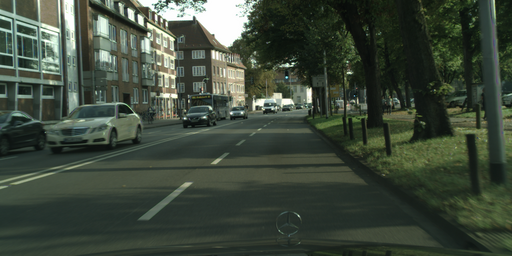} & \includegraphics[width=\linewidth, frame]{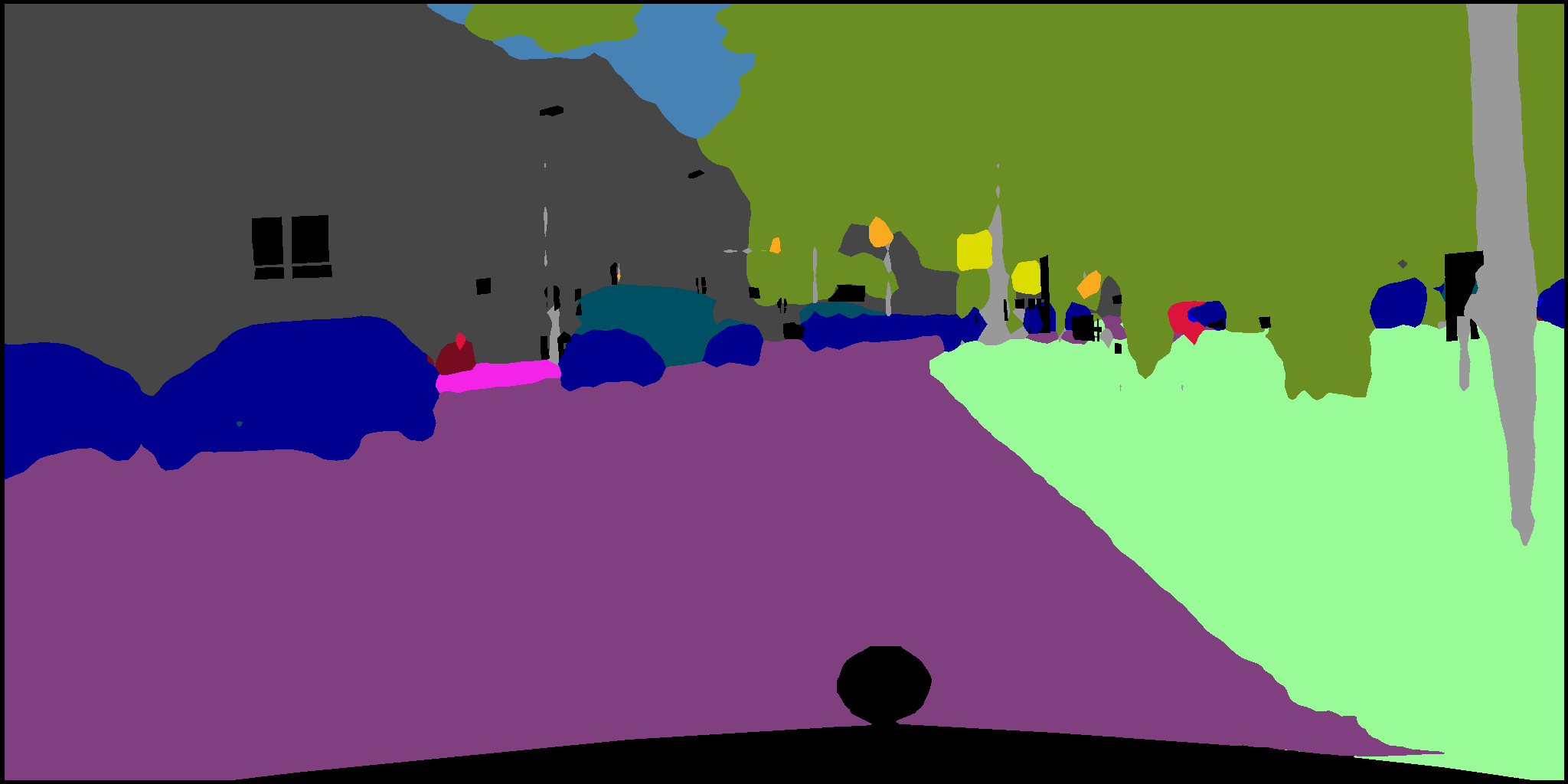} & \includegraphics[width=\linewidth, frame]{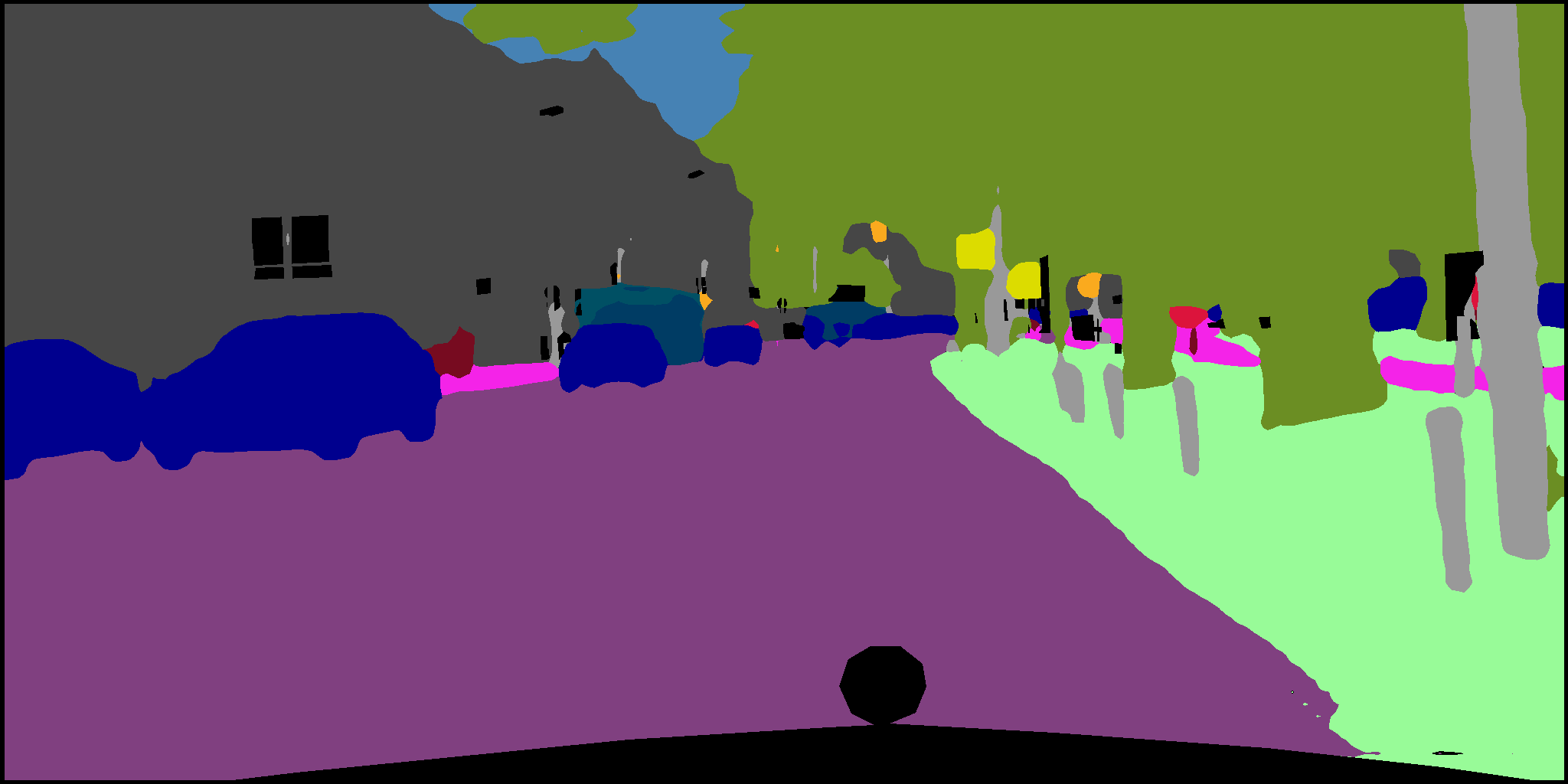} & \includegraphics[width=\linewidth, frame]{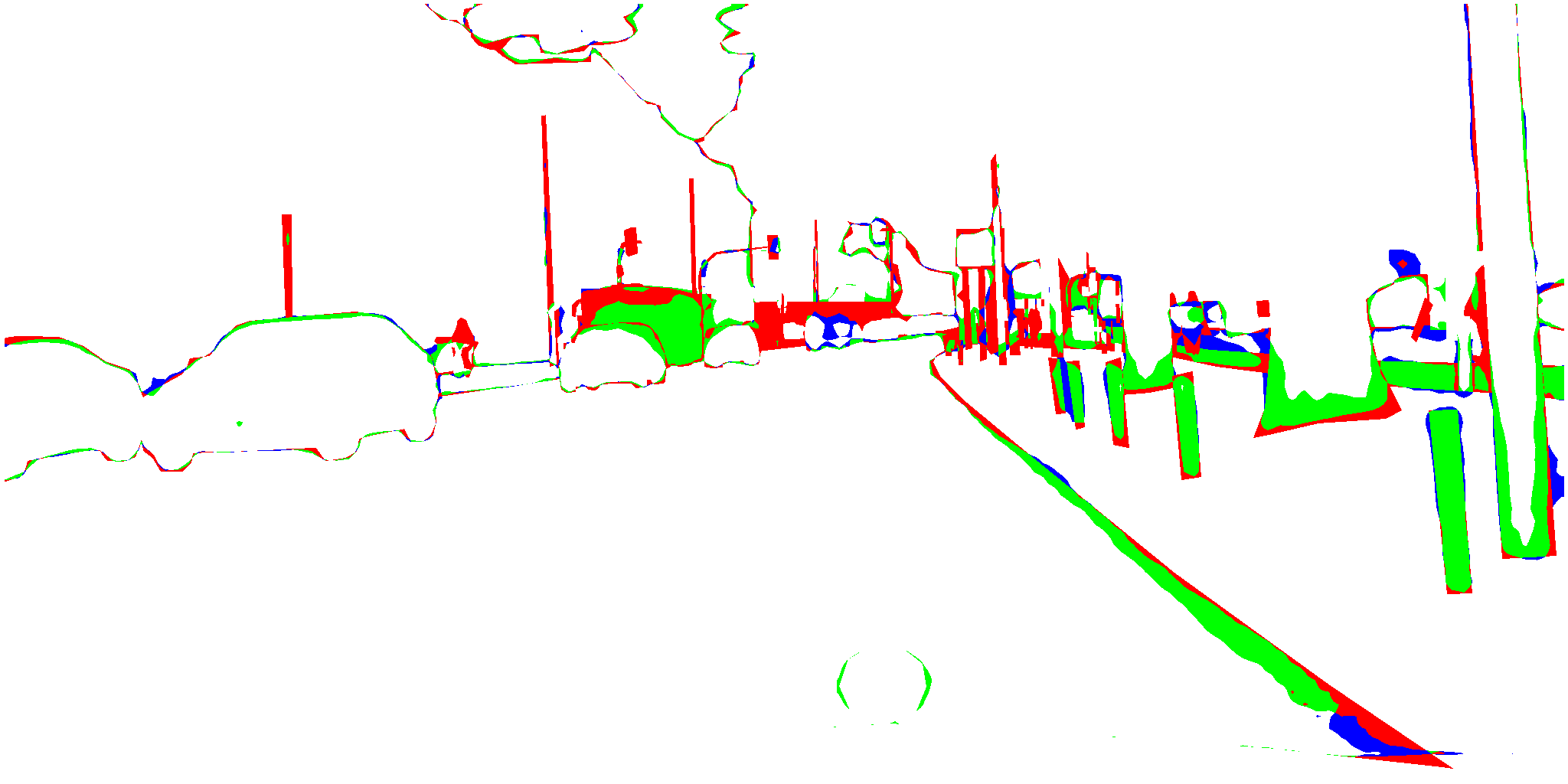}\\
\\
(iv) & \includegraphics[width=\linewidth,  frame]{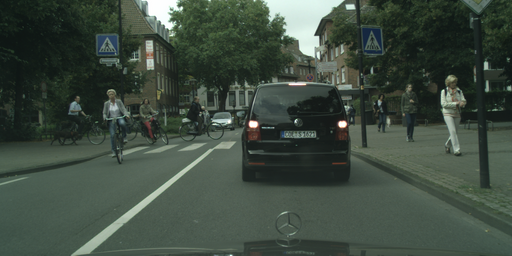} & \includegraphics[width=\linewidth, frame]{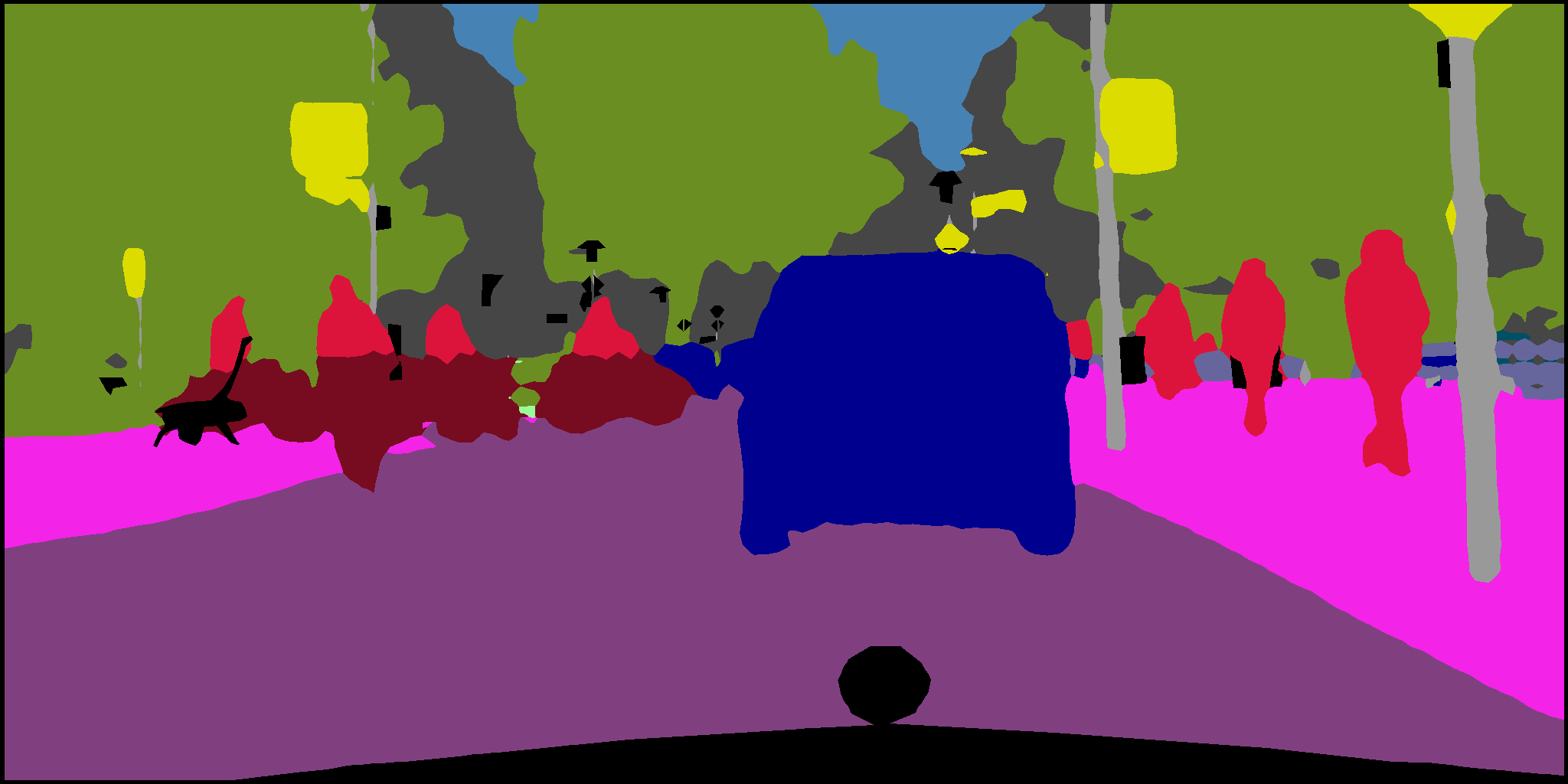} & \includegraphics[width=\linewidth, frame]{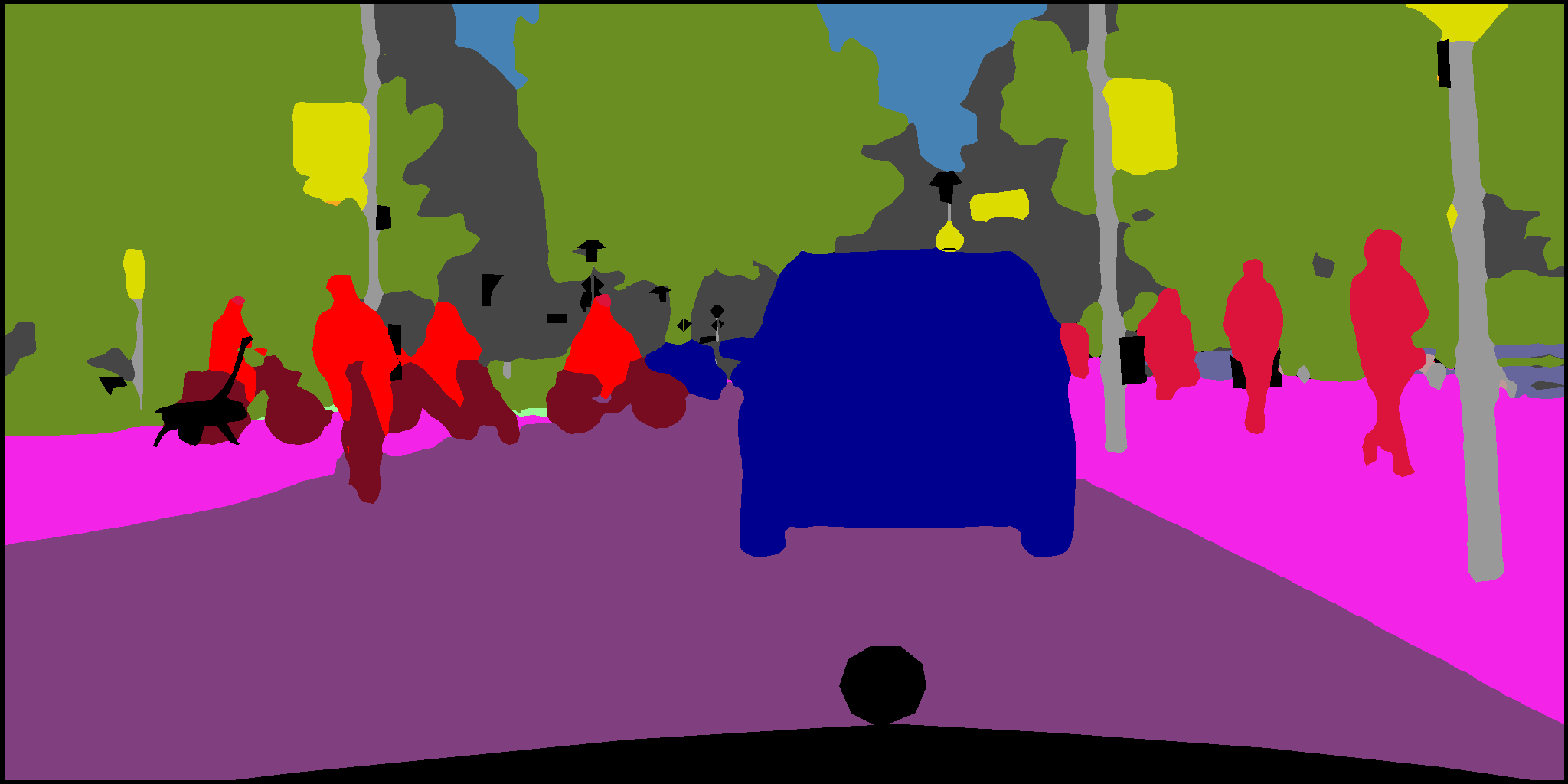} & \includegraphics[width=\linewidth, frame]{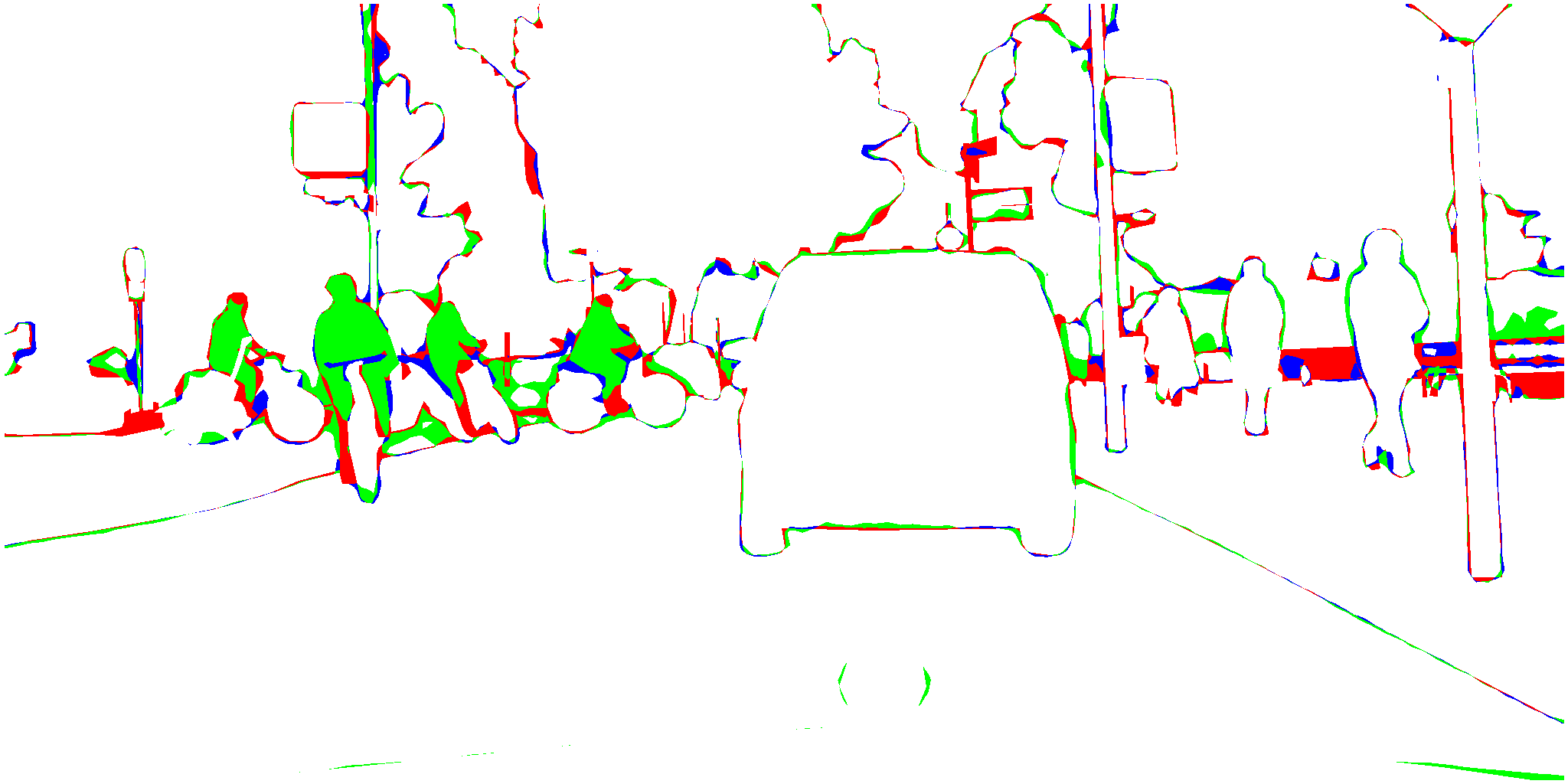}\\
\end{tabular}
}
\caption{Semantic segmentation results on Cityscapes 10-1 (10 tasks).}
\label{fig:qual-analysisIS}
\end{subfigure}
\hfill
\begin{subfigure}[b]{0.495\textwidth}
\vskip 0pt
\centering
\footnotesize
\setlength{\tabcolsep}{0.05cm}% for the horiz padding
{
\renewcommand{\arraystretch}{0.2}% for the vertical padding
\newcolumntype{M}[1]{>{\centering\arraybackslash}m{#1}}
\begin{tabular}{cM{1.95cm}M{1.95cm}M{1.95cm}M{1.95cm}}
& Input image & DKD~\cite{baek2022_dkd} & \net~(Ours) & Improv./Error \\
(i) & \includegraphics[width=\linewidth, frame, trim={0 3.5cm 0 3.5cm},clip]{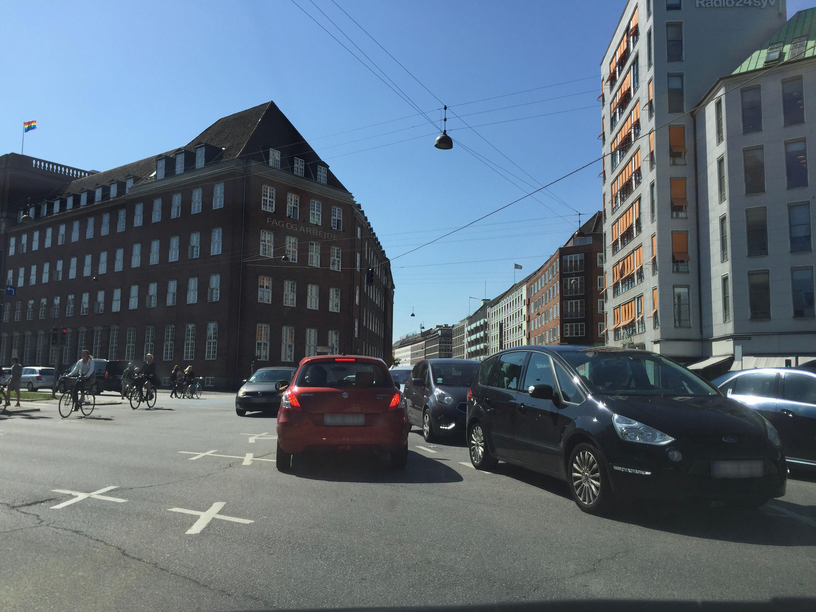} & \includegraphics[width=\linewidth, frame, trim={0 14cm 0 14cm},clip]{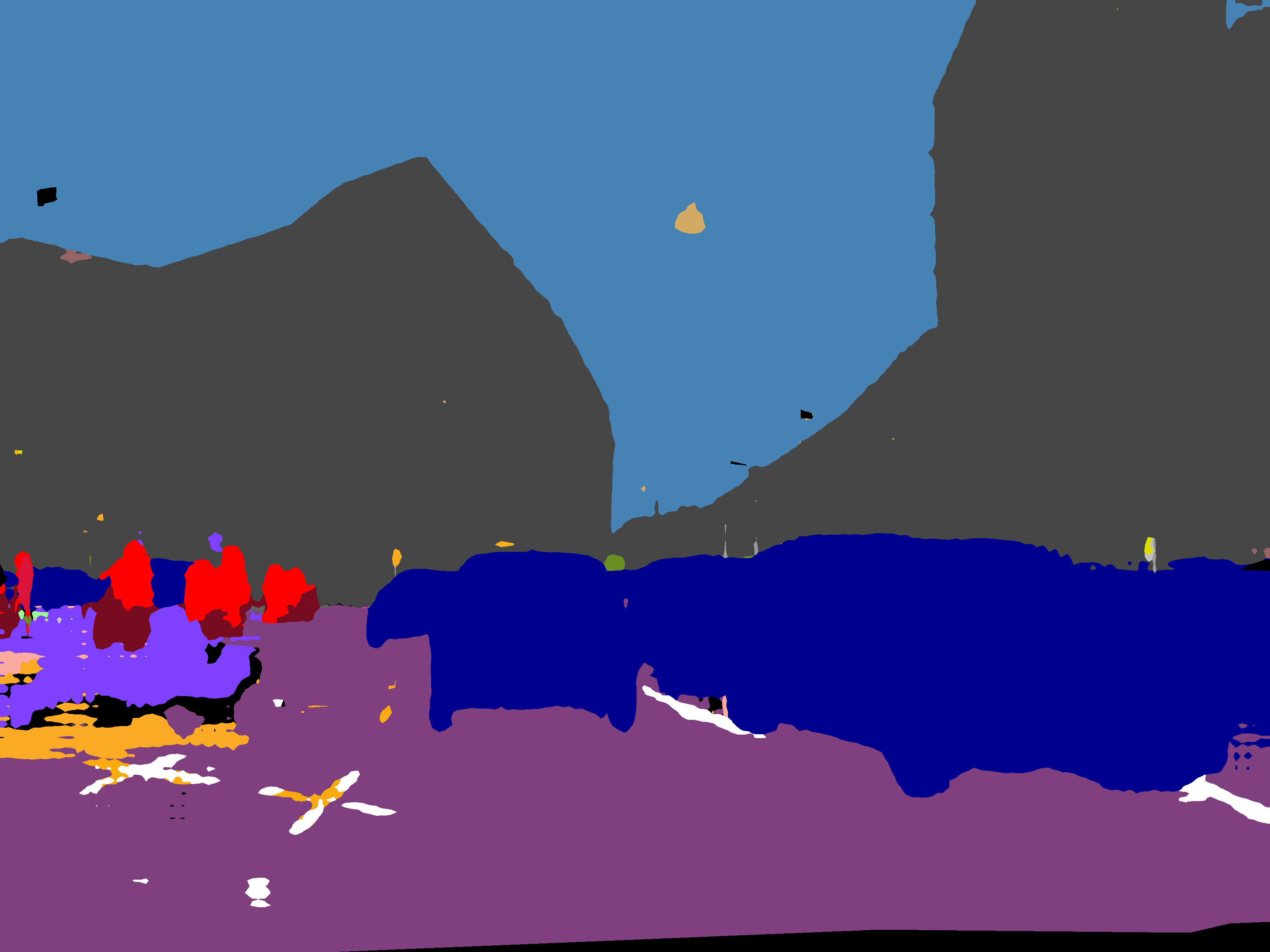} & \includegraphics[width=\linewidth, frame, trim={0 14cm 0 14cm},clip]{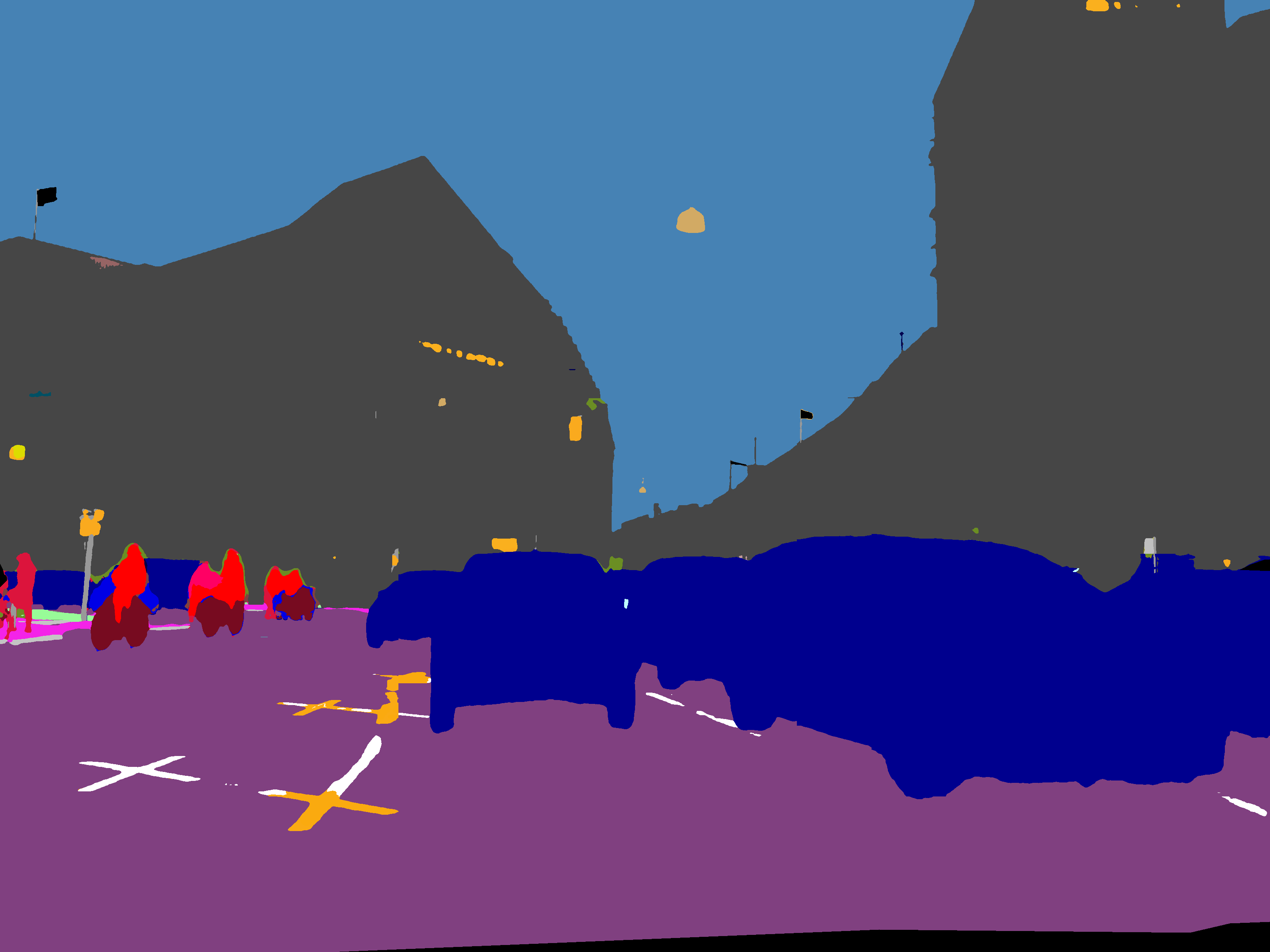} & \includegraphics[width=\linewidth, frame, trim={0 14cm 0 14cm},clip]{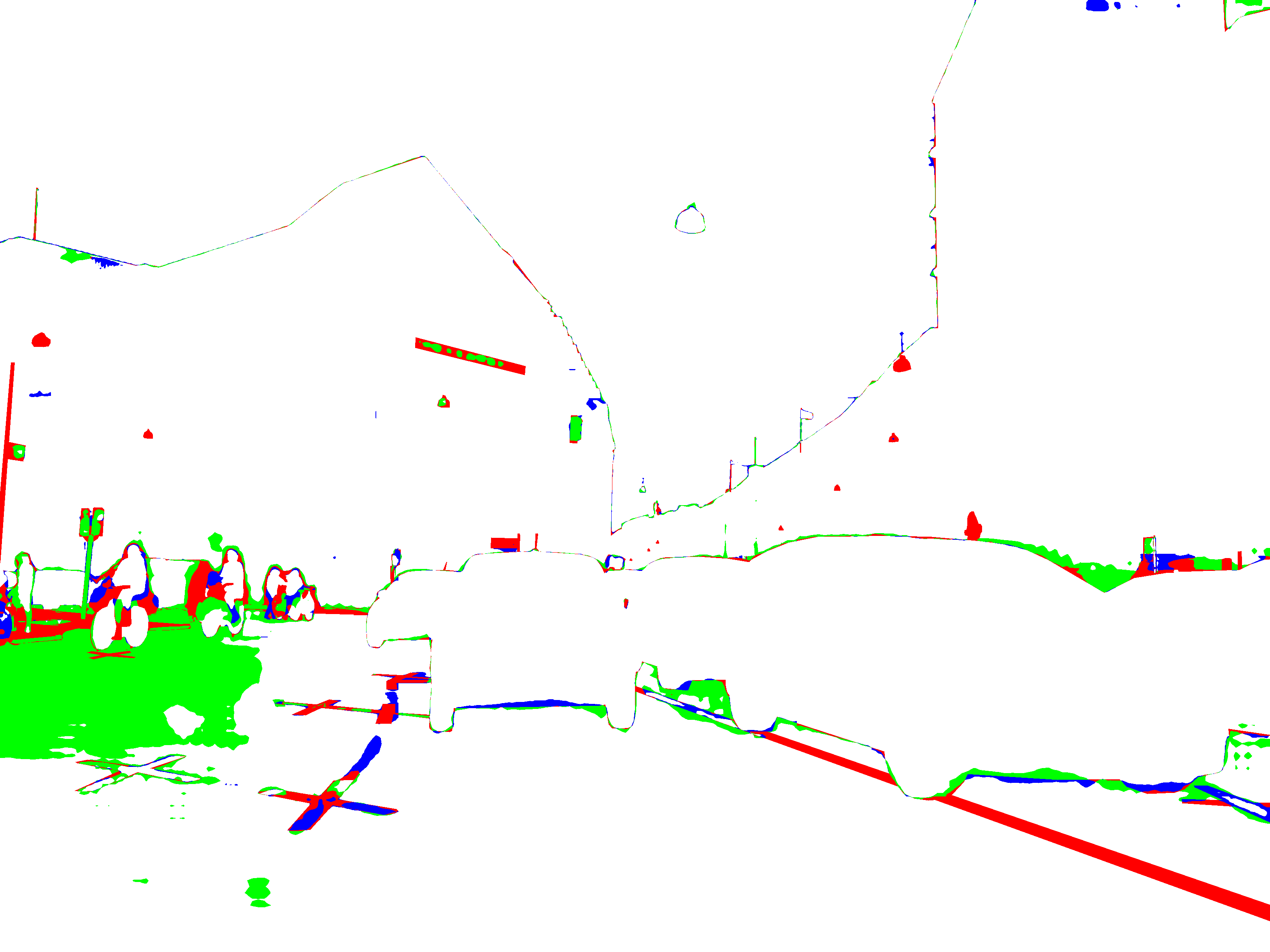}\\
\\
(ii) &  \includegraphics[width=\linewidth, frame, trim={0 3.5cm 0 3.5cm},clip]{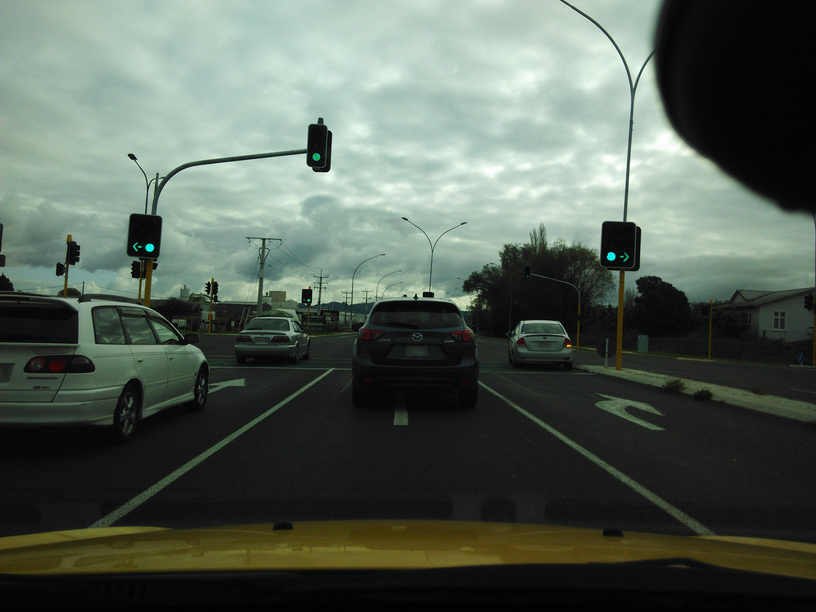} & \includegraphics[width=\linewidth, frame, trim={0 14cm 0 14cm},clip]{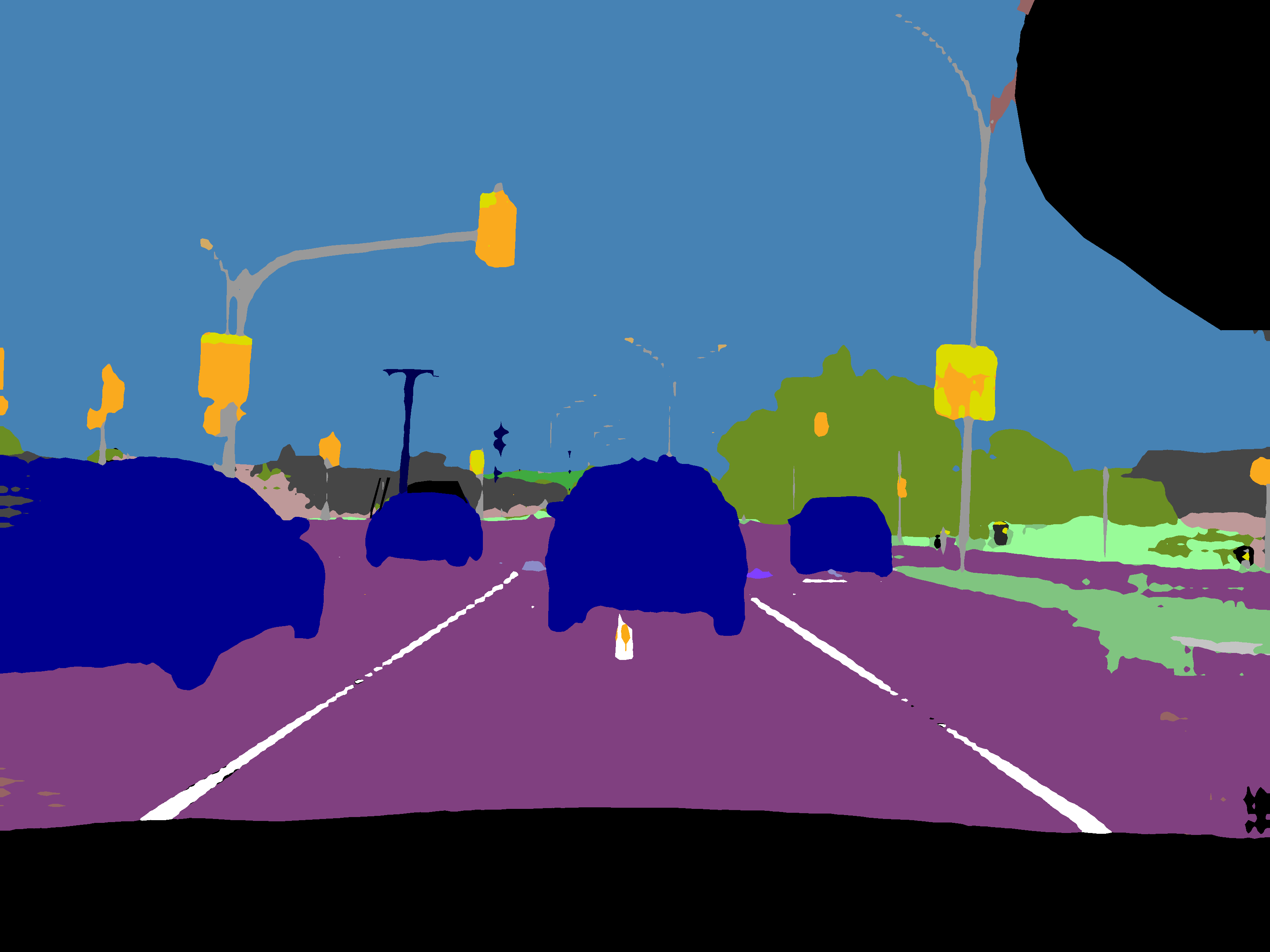} & \includegraphics[width=\linewidth, frame, trim={0 14cm 0 14cm},clip]{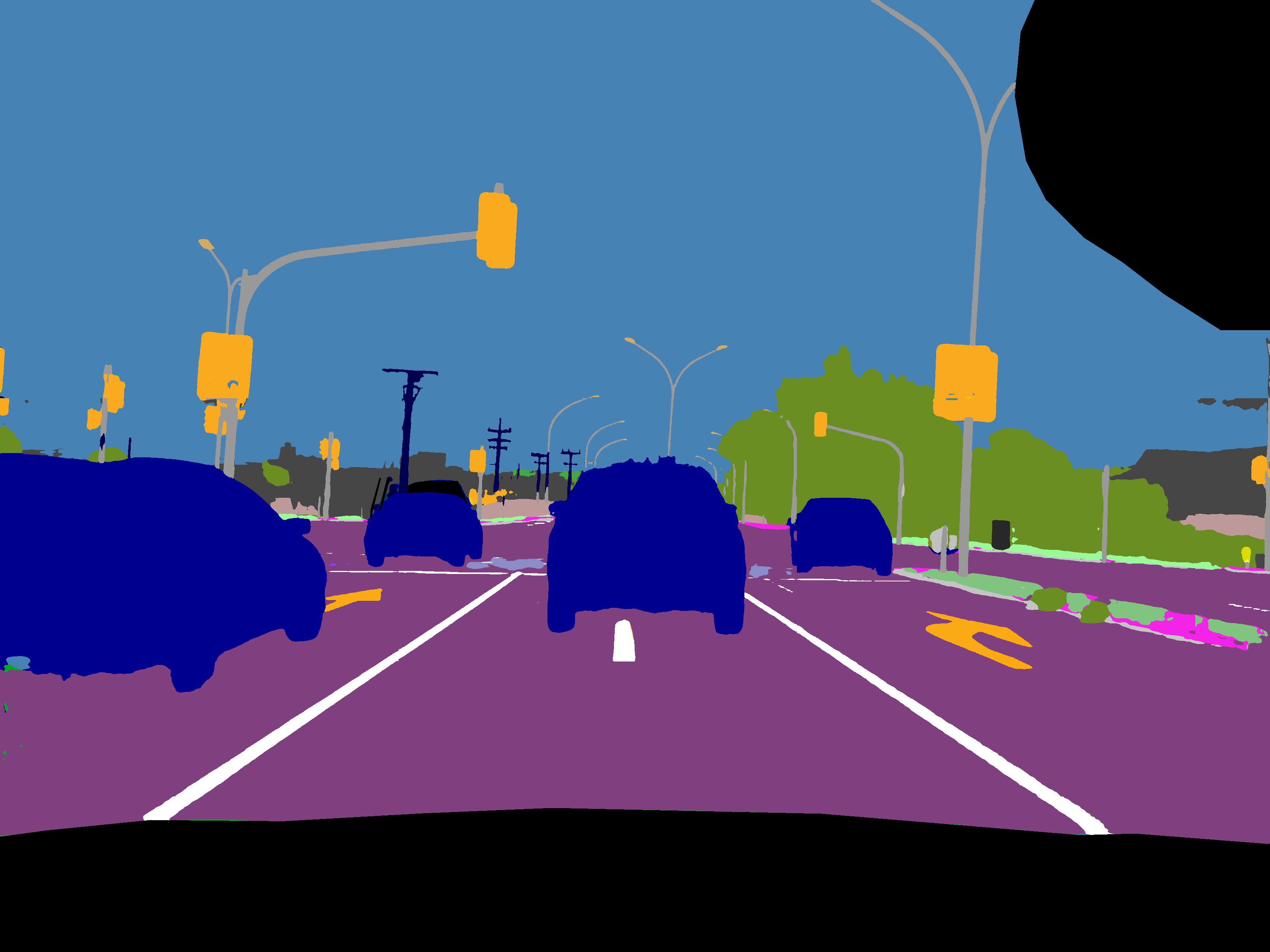} & \includegraphics[width=\linewidth, frame, trim={0 14cm 0 14cm},clip]{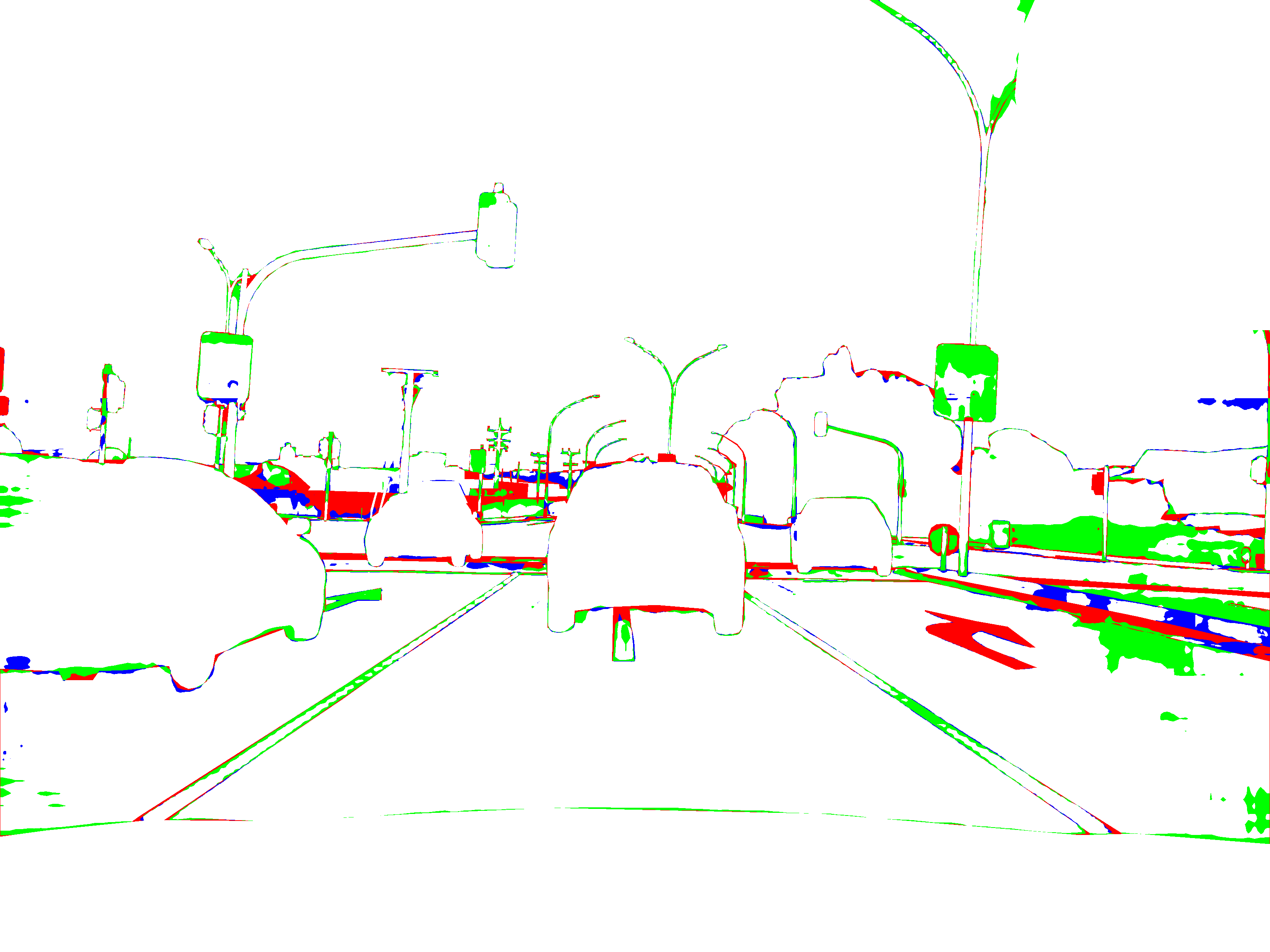}\\
\\
(iii) & \includegraphics[width=\linewidth, frame, trim={0 2.1cm 0 2.1cm},clip]{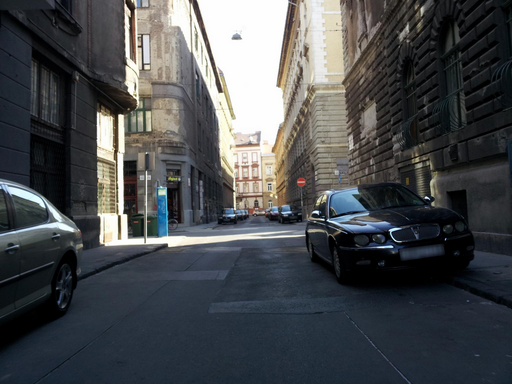} & \includegraphics[width=\linewidth, frame, trim={0 8.5cm 0 8.5cm},clip]{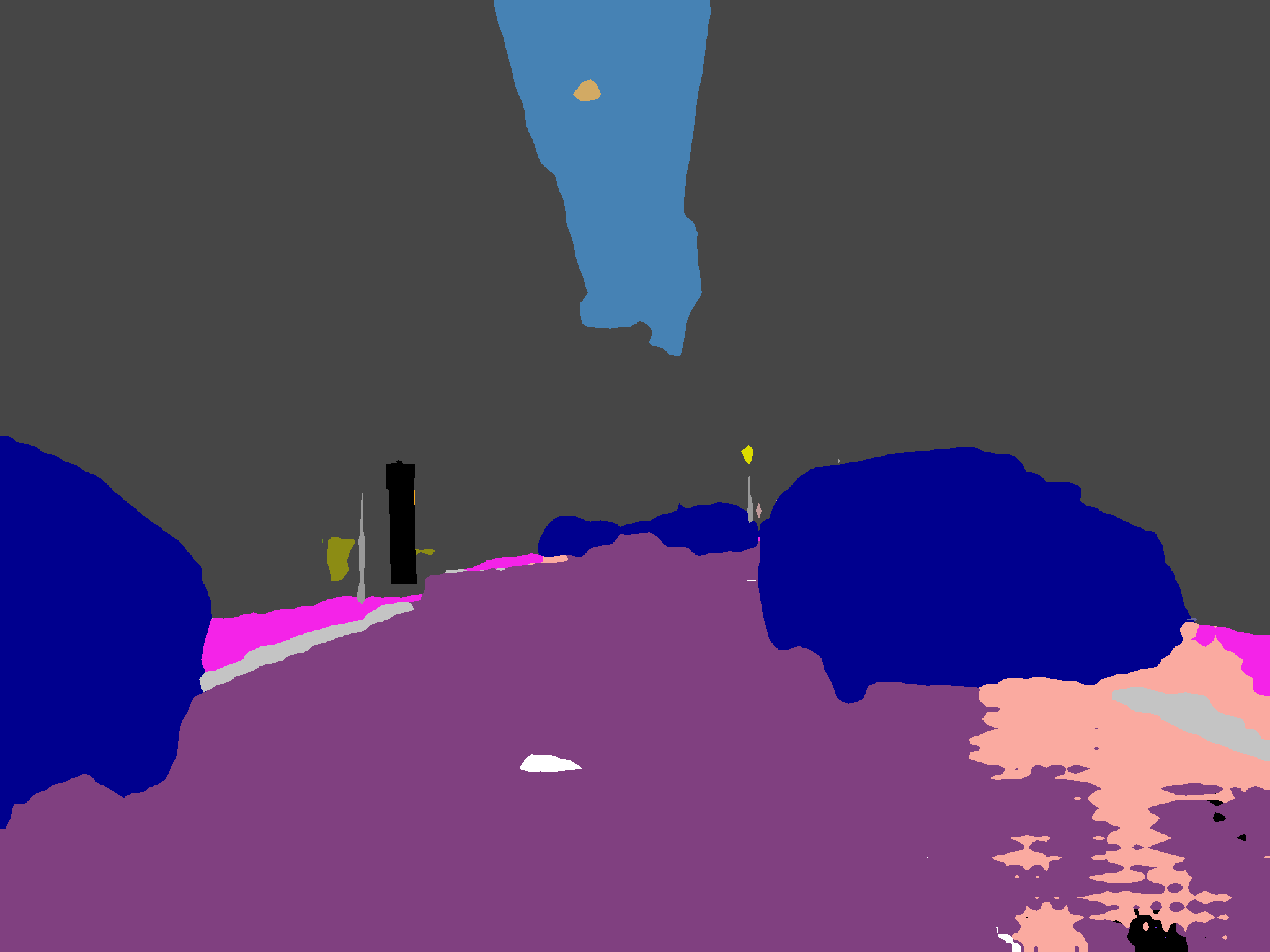} & \includegraphics[width=\linewidth, frame, trim={0 8.5cm 0 8.5cm},clip]{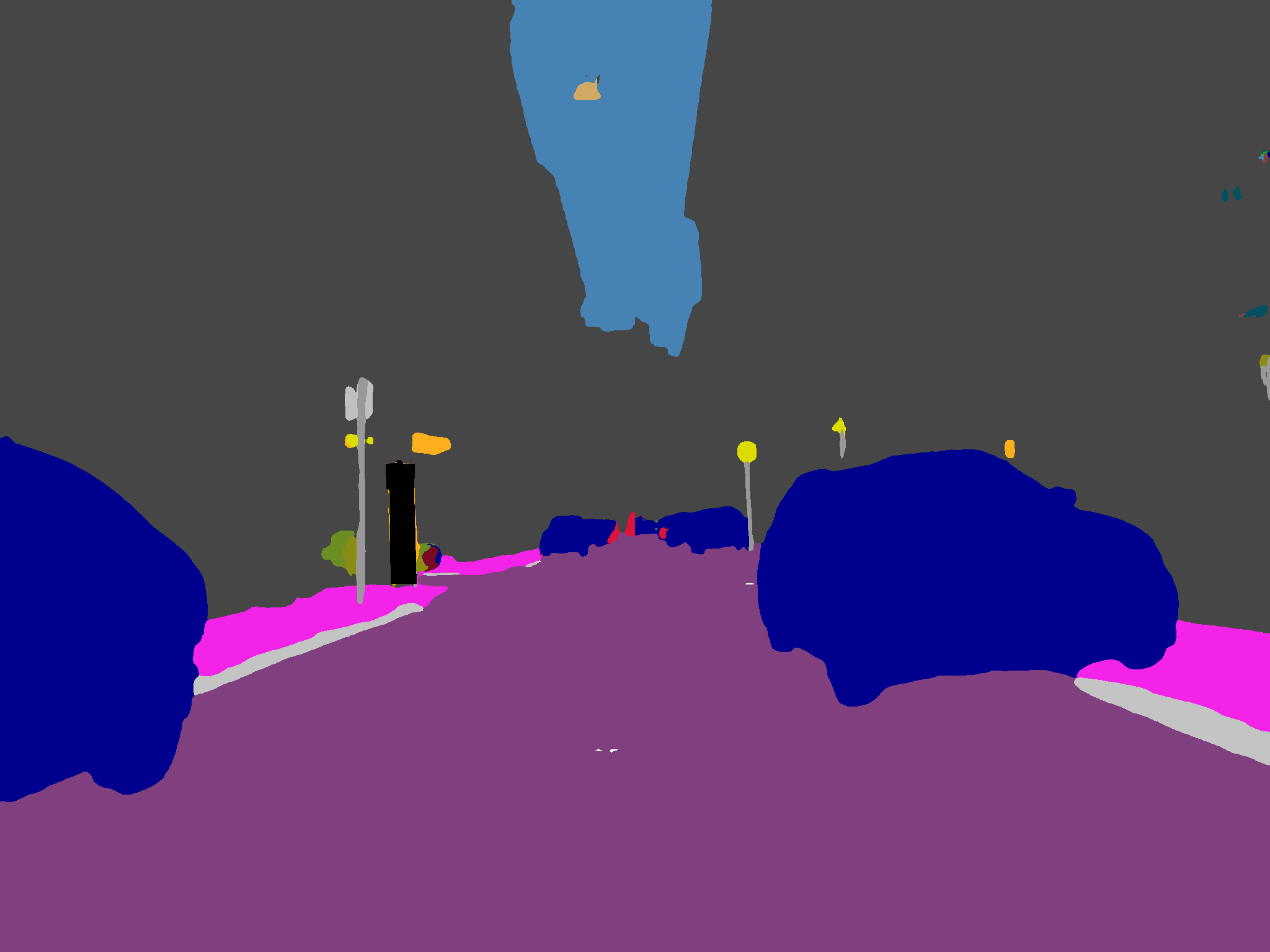} & \includegraphics[width=\linewidth, frame, trim={0 14cm 0 14cm},clip]{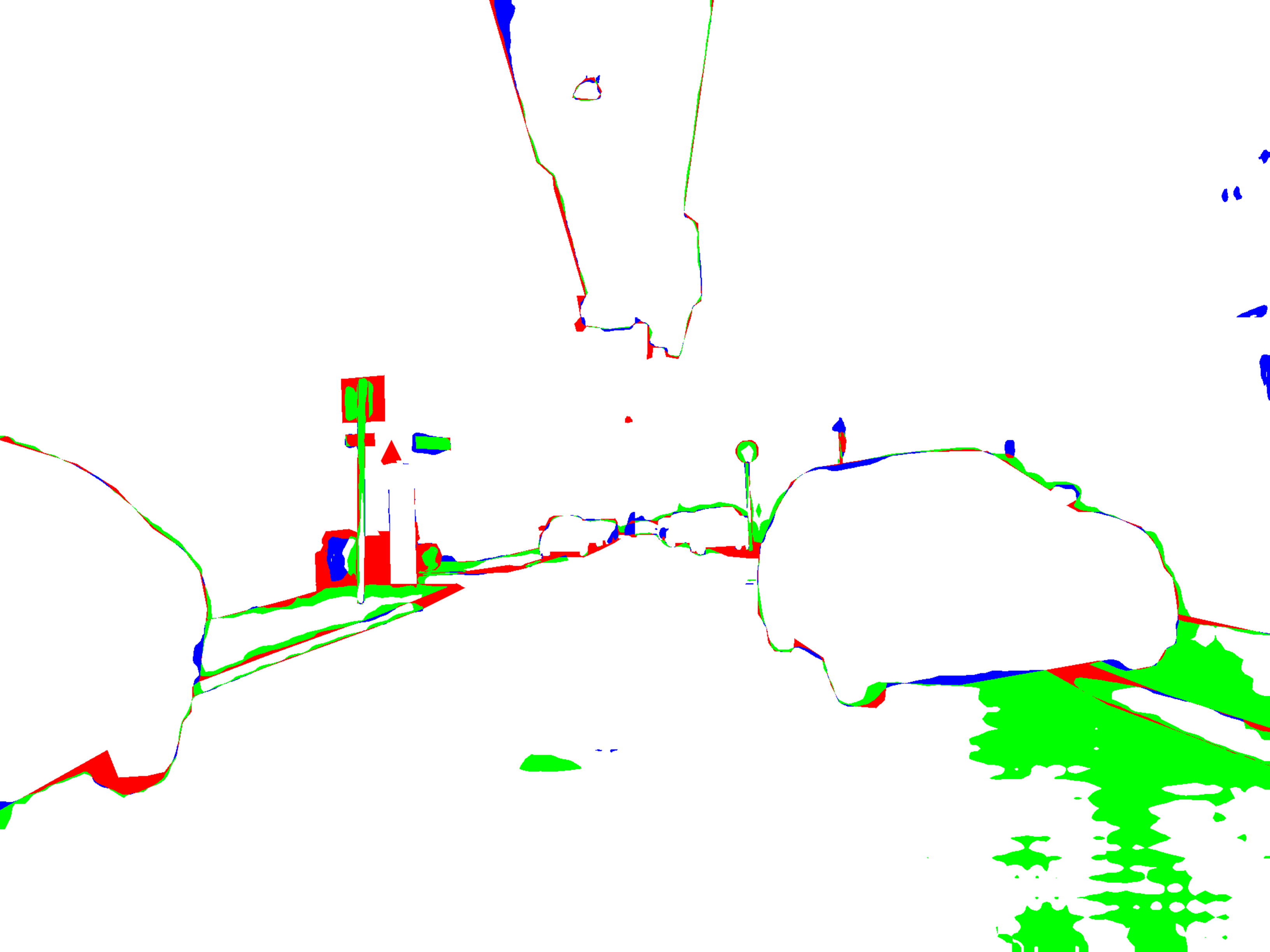}\\
\\
(iv) & \includegraphics[width=\linewidth,  frame, trim={0 3.5cm 0 3.5cm},clip]{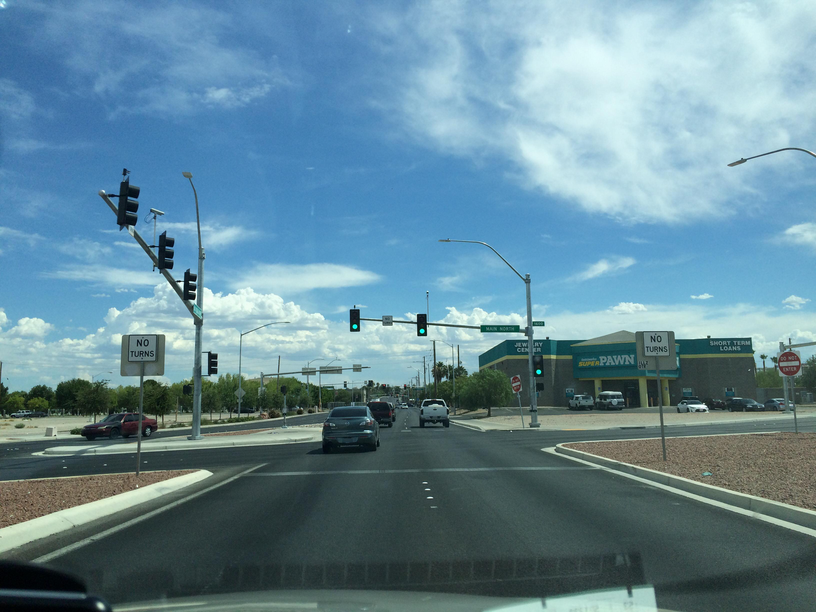} & \includegraphics[width=\linewidth, frame, trim={0 14cm 0 14cm},clip]{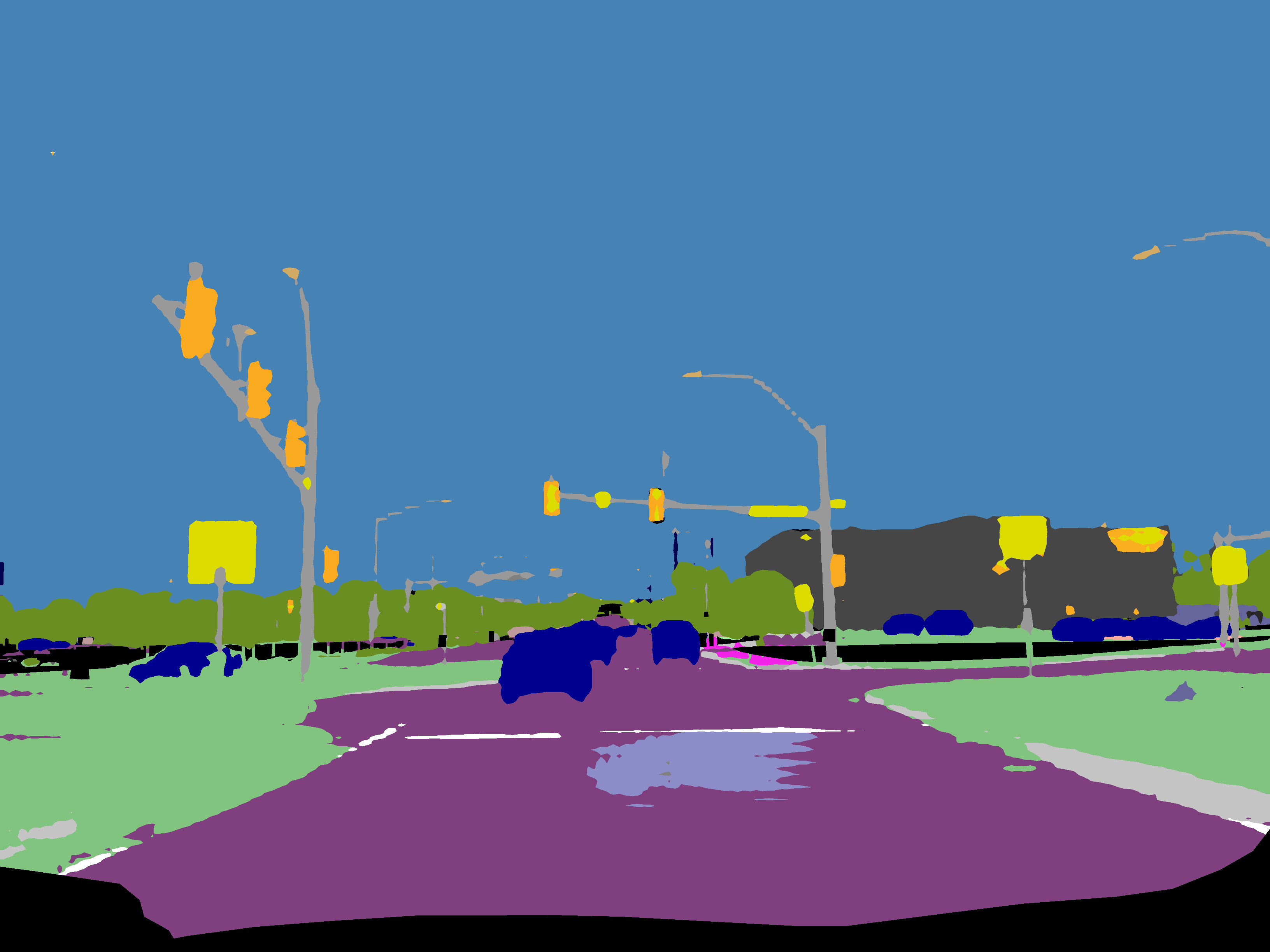} & \includegraphics[width=\linewidth, frame, trim={0 14cm 0 14cm},clip]{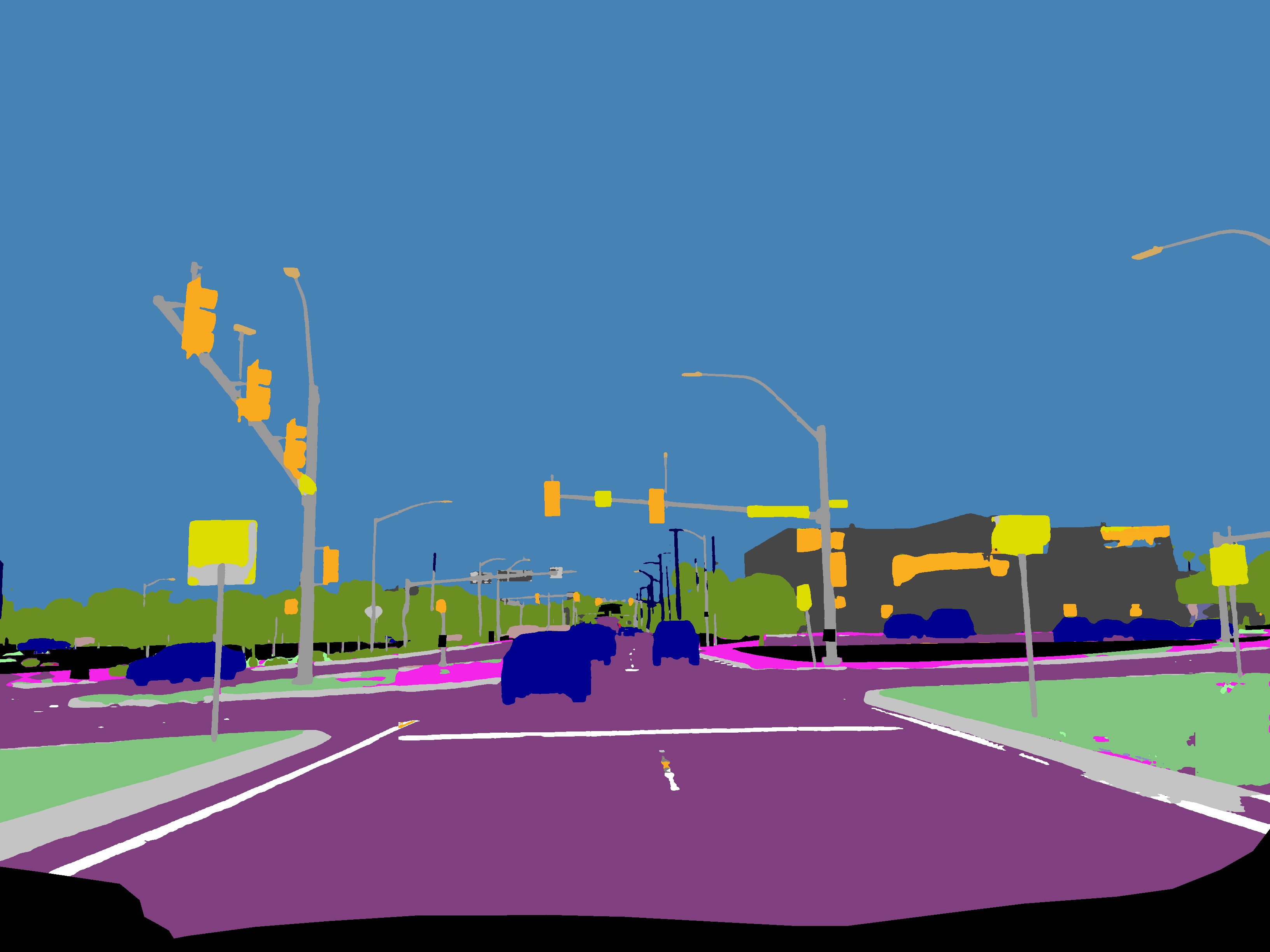} & \includegraphics[width=\linewidth, frame, trim={0 14cm 0 14cm},clip]{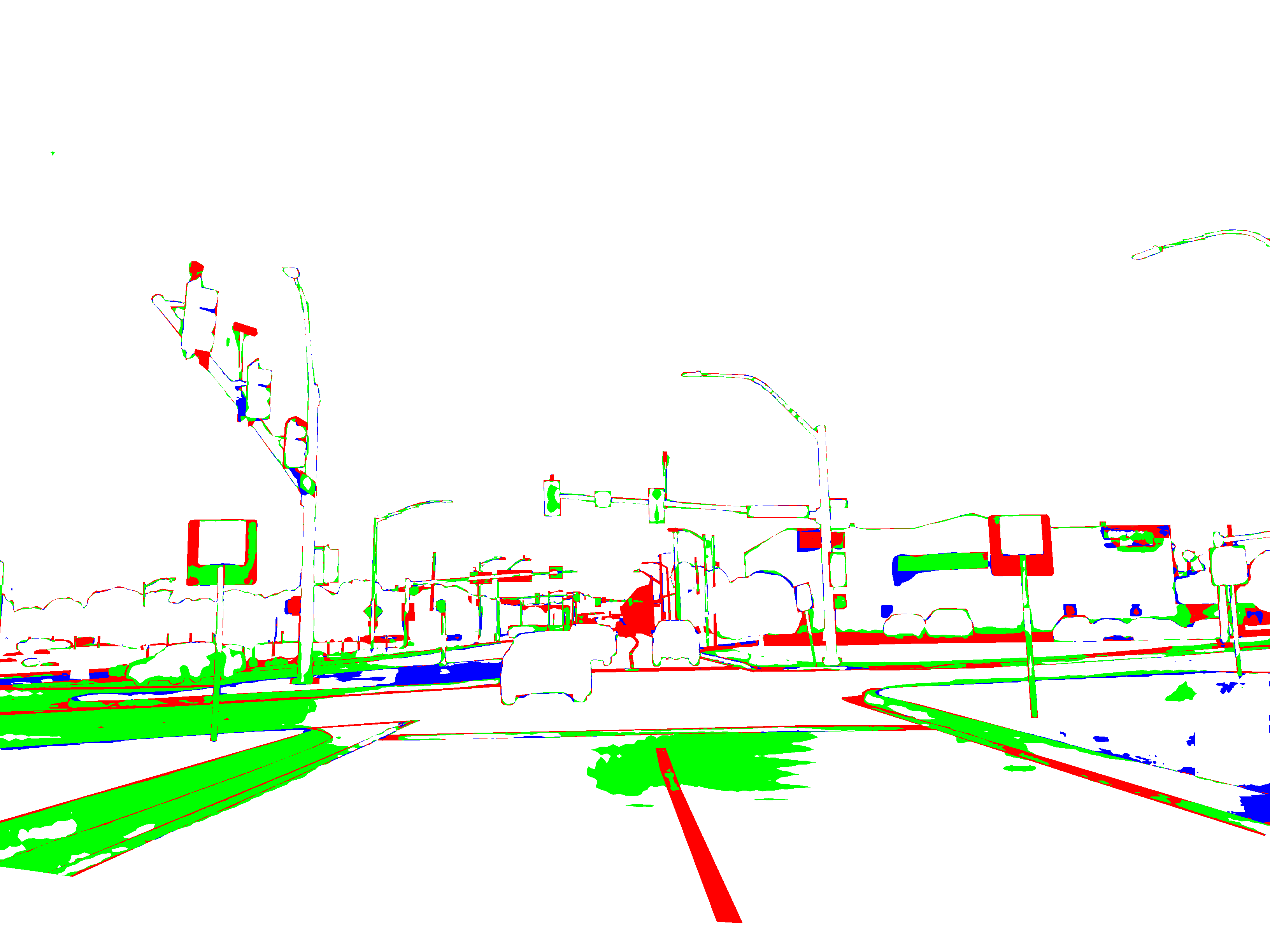}\\
\end{tabular}
}
\caption{Semantic segmentation results on Mapillary Vistas 2.0 71-10 (5tasks).}
\label{fig:qual-analysisSS}
\end{subfigure}
\caption{Qualitative results of \net~in comparison to the best-performing baselines for CISS on the Cityscapes and Mapillary Vistas 2.0 dataset. The improvement/error maps show pixels misclassified by the baseline and correctly predicted by \net~in green and vice-versa in blue. Incorrect predictions of both models are colored in red.}
\vspace{-0.5cm}
\end{figure*}

\subsubsection{Weighting of Loss Functions} \label{sec:ds}
We ablate the weighting of our hierarchical relation distillation loss and hyperbolic distance correlation loss in~\tabref{tab:reg}. We note that a higher weighting of either loss increases the rigidity of the model which can be observed as higher base class scores. As a result, the plasticity is negatively affected which is evident through lower novel class performances for the relation distillation loss. Consequently, we find that the weighting can be used to tailor the desired plasticity and rigidity of an incremental model.\looseness=-1

\begin{table}
\centering
\caption{Influence of regularisation weighting term for results on Cityscapes in mIoU (\%).}
\label{tab:reg}
\begin{tabular}
{ll|ccc}
 \toprule
\multicolumn{5}{c}{\textbf{14-1 (6 tasks)}} \\
\textbf{$\mathcal{L}_{dist}$} & \textbf{$\mathcal{L}_{rel}$} & 1-14 & 15-19 & all \\
\midrule 
0.001 & 10 & 72.92 & 42.64 & 61.70 \\
0.1 & 10 & \textbf{73.10} & 42.21 & 61.72 \\
0.01 & 10 & 73.03 & \textbf{42.47} & \textbf{61.74}\\
0.01 & 1 & 73.05 & 42.08 & 61.65 \\
0.01 & 100 & 73.08 & 41.29 & 61.48 \\
\hline
\end{tabular}
\vspace{-0.3cm}
\end{table}

\subsubsection{Taxonomic Hierarchy and Hierarchical Loss} \label{sec:hh}
We present the performance of different hierarchical loss functions in~\tabref{tab:hr}. \rebuttal{In incremental steps, the tree-min loss significantly outperforms simple hierarchical BCE and CE losses where the logits of leaf nodes $v_L$ are multiplied with its ancestors scores ($\mathcal{A}_{v_L}$)}. Further, we note that the \rebuttal{distance-based margin loss introduced in~\cite{liulei24} does not result in a performance improvement in hyperbolic space. We reason that constant margin-based losses are not effective with non-linear distance changes in hyperbolic space.} Further, we ablate different taxonomy trees (shown in the supplementary material \ifdefined\SUP\secref{sec:illu}\else Sec. \textcolor{red}{S.2}\fi) in~\tabref{tab:tree}. We observe a significant performance decrease of \rebuttal{$10.99$pp} when formulating the incremental task with only \rebuttal{two task levels}. Consequently, we emphasize the benefit of deep hierarchies \rebuttal{in base and incremental training. More results are in the supplementary material \ifdefined \SUP \secref{sec:sup_h}\else Sec. \textcolor{red}{S.3}\fi, \ifdefined \SUP \secref{sec:sup_tt} \else Sec. \textcolor{red}{S.4} \fi and \ifdefined \SUP \secref{sec:sup_m}\else Sec. \textcolor{red}{S.6}\fi}.

\begin{table}
\centering
\caption{Influence of different hierarchical loss functions for results on Cityscapes in mIoU (\%).}
\label{tab:hr}
\begin{tabular}
{l|cccc}
 \toprule
\multicolumn{5}{c}{\textbf{14-1 (6 tasks)}} \\
\textbf{Hierarchical Loss Fct.} & 1-14$_{0}$ & 1-14 & 15-19 & all \\
\midrule 
Hierarchical BCE & \rebuttal{66.87} & 60.72 & 22.06 & 48.02\\
\rebuttal{Hierarchical CE~\cite{atigh2022hyperbolic}} & \rebuttal{73.61} & \rebuttal{62.89} & \rebuttal{33.92} & \rebuttal{52.51}\\
Tree-min & \rebuttal{\textbf{74.85}} & \textbf{73.03} & \textbf{42.47} & \textbf{61.74}\\
Tree-min + \ Margin loss~\cite{liulei24} & \rebuttal{74.20} & 72.97 & 
35.31 & 59.91 \\
\hline
\end{tabular}
\vspace{-0.2cm}
\end{table}

\begin{table}[t]
\centering
\caption{Influence of different class taxonomy trees ($\mathcal{H}$) for results on Cityscapes in mIoU (\%).}
\label{tab:tree}
\begin{tabular}
{l|ccccc}
 \toprule
\multicolumn{5}{c}{\textbf{14-1 (6 tasks)}} \\
\textbf{$\mathcal{H}$} & 1-14$_{0}$ &1-14 & 15-19 & all \\
\midrule
$\mathcal{H}_2$ & \rebuttal{66.46} & 60.86 & 32.57 & 50.75 \\
$\mathcal{H}_4$ & \rebuttal{74.54} & 72.51 & 30.44 & 58.37 \\
$\mathcal{H}_6$ & \rebuttal{\textbf{74.85}} & \textbf{73.03} & \textbf{42.47} & \textbf{61.74} \\
\hline
\end{tabular}
\vspace{-0.3cm}
\end{table}

\subsubsection{Curvature} \label{sec:curv}
We observe the influence of different curvatures in~\tabref{tab:curv}. \rebuttal{While we and \cite{atigh2022hyperbolic} report minor performance changes as an effect of this hyperparameter in closed-set training, we find it has a significant impact on our CISS task.} While a lower curvature ($c=0.1$) eases learning novel classes, the forgetting of base classes is more pronounced. Further, a high curvature ($c=10$) results in worse performance for all classes. We reason this observation with the reduced size of the feature space.\looseness=-1

\begin{table}[t]
\centering
\caption{Influence of different curvatures for results on Cityscapes in mIoU (\%).}
\label{tab:curv}
\begin{tabular}
{l|cccc}
 \toprule
\multicolumn{5}{c}{\textbf{14-1 (6 tasks)}} \\
\textbf{Curvature} & 1-14$_{0}$ &1-14 & 15-19 & all \\
\midrule
0.1 & 74.24 & 71.82 & 42.24 & 60.84 \\
1 & \rebuttal{74.48} & 72.96 & 38.33 & 60.66\\
2 & \textbf{74.85} & 73.03 & \textbf{42.47} & \textbf{61.74} \\
5 & 74.59 & \textbf{73.18} & 37.39 & 60.57\\
10 & 74.77 & 72.93 & 35.56 & 59.94 \\
\hline
\end{tabular}
\vspace{-0.3cm}
\end{table}

\subsection{Qualitative Results}
We present qualitative evaluations of \net~with the best-performing baseline on Cityscapes in~\figref{fig:qual-analysisIS}. We observe that both methods can precisely segment various classes e.g., pedestrians, cars, and vegetation. While DKD~\cite{baek2022_dkd} tends to incorrectly classify flat surfaces as background in~\figref{fig:qual-analysisIS}~(i, ii), our method superiorly remembers those old classes and continuous to accurately predict them after having learned new classes. This characteristic can be attributed to our strategy of maintaining knowledge by modeling the class hierarchy in hyperbolic space.
Further, we note that DKD~\cite{baek2022_dkd} merges objects with their respective background in \figref{fig:qual-analysisIS}~iii) and is not able to distinguish between the incremental classes 'rider' and 'pedestrian' which are learned sequentially. On the other hand, our method rarely confuses the named classes and maintains knowledge of all incremental classes. We reason this observation with our knowledge retention losses which ensures that relevant relations are maintained when new classes are added.\looseness=-1

On Mapillary Vistas v2.0, we observe that our approach and the best baseline can superiorly detect 'cars' and 'traffic lights' which are learned in the last two incremental steps as shown in~\figref{fig:qual-analysisSS}. However, while \net~is consistently able to identify the complete road structure, DKD~\cite{baek2022_dkd} incorrectly predicts large shapes of other classes in~\figref{fig:qual-analysisSS}~(i, iii, iv). We find that this behavior is hazardous in autonomous driving scenarios where misdetections lead to emergency braking. Both methods fail to predict markings and capture the complete shape of traffic signs in \figref{fig:qual-analysisSS}~(i) and \figref{fig:qual-analysisSS}~(iv). %We believe that a stronger model can resolve this problem and highlight that \net~can be readily employed with different backbones.

\section{Conclusion}
\label{sec:conclusion}

We present \net, a novel CISS approach that models features conforming to taxonomy-tree structures on the Poincaré ball to balance rigidity and plasticity in incremental learning. \net~further maintains implicit class relations between old class hyperplanes and constraints features to have equidistant radii. We presented extensive experimental evaluations of eight incremental settings on Cityscapes and Mapillary Vistas~2.0 that demonstrated that \net~achieves state-of-the-art performance. Our method is one of the early works that uniformly addresses the bifurcation of previously observed classes and incremental classes from the background. Further, we emphasize the benefit of hierarchical modeling in hyperbolic space and motivate future work to explore its potential for various open-world challenges.

%\section*{Acknowledgement}
%The authors thank Kshitij Sirohi for technical discussions.

%%%%%%%%%%%%%%%%%%%%%%%%%%%%%%%%%%%%%%%%%%%%%%%%%%%%%%%%%%%%%%%%%%%%%%%%%%%%%%%%

{\footnotesize
\bibliographystyle{IEEEtran}
\bibliography{references}
}

\ifdefined\SUP
    %%%%%%%%%% Merge with supplemental materials %%%%%%%%%%
\clearpage
\renewcommand{\baselinestretch}{1}

\begin{strip}
\begin{center}
\vspace{-5ex}

\textbf{\Large \bf
Taxonomy-Aware Continual Semantic Segmentation in\\Hyperbolic Spaces for Open-World Perception
} \\
\vspace{3ex}

\Large{\bf- Supplementary Material -}\\
\vspace{0.4cm}
\normalsize{Julia Hindel, Daniele Cattaneo and Abhinav Valada}
\end{center}
\end{strip}

%%%%%%%%%% Merge with supplemental materials %%%%%%%%%%
%%%%%%%%%% Prefix a "S" to all equations, figures, tables and reset the counter %%%%%%%%%%
\setcounter{section}{0}
\setcounter{equation}{0}
\setcounter{figure}{0}
\setcounter{table}{0}
\makeatletter

\renewcommand{\thesection}{S.\arabic{section}}
\renewcommand{\thesubsection}{S.\arabic{section}.\Alph{subsection}}
\renewcommand{\thetable}{S.\arabic{table}}
\renewcommand{\thefigure}{S.\arabic{figure}}

\let\thefootnote\relax\footnote{ Department of Computer Science, University of Freiburg, Germany.}%
\normalsize

\normalsize
\begin{table}[t]
\centering
\caption{Influence of different class taxonomy trees ($\mathcal{H}$) for results on Cityscapes in mIoU (\%). The setting $1-14_0$ represents the mIoU on base classes at time $t=0$. The relative decrease in performance on $\mathcal{C}^1$ is recorded in (\%).}
\label{tab:tree_sup}
\begin{tabular}
{l|ccccc}
 \toprule
 \multicolumn{6}{c}{\textbf{14-1 (6 tasks)}} \\
\textbf{$\mathcal{H}$} & 1-14$_{0}$ &$\downarrow$1-14$_{0}$ [\%] & 1-14 &  15-19 & all \\
\midrule
$\mathcal{H}_3a$ & 69.05 & 4.10 & 66.22 & 29.17 & 56.95 \\
$\mathcal{H}_4a$ & 69.61 & 2.99 & 67.53 &34.30 & 59.22\\
$\mathcal{H}_5a$ & 69.35 & 3.03 & 67.25 & 30.59 & 58.08\\
$\mathcal{H}_6$ & \textbf{74.84} & \textbf{2.42} & \textbf{73.03} & \textbf{42.47} & \textbf{61.74} \\
\hline
\end{tabular}
%\vspace{-0.3cm}
\end{table}

\begin{table}[t]
\centering
\caption{Influence of different class taxonomy trees ($\mathcal{H}$) for computational cost on Cityscapes in mIoU (\%).}
\label{tab:time_rebut}
\begin{tabular}
{l|cc}
 \toprule
\multicolumn{3}{c}{\textbf{14-1 (6 tasks)}} \\
\textbf{$\mathcal{H}$} & training time on GPU & average inference time in seconds\\
\midrule
$\mathcal{H}_2$ & 4 hours 43 minutes & 0.22 \\
$\mathcal{H}_4$ & 6 hours 49 minutes & 0.22 \\
$\mathcal{H}_6$ & 8 hours 16 minutes & 0.24 \\
\hline
\end{tabular}
%\vspace{-0.3cm}
\end{table}

\begin{table*}
\centering
\caption{Continual semantic segmentation results on Cityscapes in mIoU (\%). Tasks defined as $\mathcal{C}^1$-$\mathcal{C}^T$($T$ tasks) and $h$ class hierarchy increments. The relative decrease in performance on $\mathcal{C}^1$ is recorded in (\%).}
\label{tab:cityap}
\setlength{\tabcolsep}{4pt}
\begin{tabular}
{l|lcc|lcc|lcc|lcc}
 \toprule
 & \multicolumn{3}{c|}{\textbf{14-1 (6 tasks)}} & \multicolumn{3}{c|}{\textbf{10-1 (10 tasks)}} & \multicolumn{3}{c|}{\textbf{7-4 (4 tasks)h}} & \multicolumn{3}{c}{\textbf{7-18 (2 tasks)h}}\\
  \cmidrule{2-13}
\textbf{Method} & 1-14 & $\downarrow$1-14$_{0}$ [\%] & all & 1-10 & $\downarrow$1-10$_{0}$ [\%] &  all & 1-7 & $\downarrow$1-7$_{0}$ [\%] &  all & 1-7 & $\downarrow$1-7$_{0}$ [\%] &  all\\
\midrule
PLOP~\cite{Douillard2020PLOPLW} & 63.54 & 3.64 & 48.33 & 60.75 & 5.43 & 42.96 & 88.56 & 1.25 & 20.75 & 88.73 & 1.07 & 17.99 \\
MiB~\cite{cermelli2020mib} & 66.37 & 5.75 & 50.05 & 61.80 & 9.80 & 45.73 & 77.66 & 13.0 & 9.83 & 90.10 & 0.0 & 9.64 \\
MiB + AWT~\cite{goswami2023attribution} & 65.60 & 6.84 & 50.72 & 60.97& 11.07 & 46.55 & 84.65 & 5.16 & 13.64 & 90.19 & 0.0 & 9.56 \\
DKD~\cite{baek2022_dkd} & 68.83 & 3.21 & 51.86 & 66.77 & 4.51 & 48.92 & 89.46 &0.0 & 4.98 & 89.19 & 0.1& 8.32 \\
MicroSeg~\cite{zhang22_microseg} & 51.35 & 8.86 & 38.84 & 44.37 & 15.73 & 32.78 & 86.39 & 0.0 & 5.79 & 86.37 & 0.0 & 11.26 \\
\midrule
\net~(Ours)$_{b2}$ & 64.52 & 3.21 & 54.64 &
66.35 & \textbf{0.0} & 55.83 &
88.11 & 1.0 & 46.99 & 
89.23 & 0.0 & 57.84 \\
\net~(Ours) & \textbf{73.03} & \textbf{2.43} & \textbf{61.74} & \textbf{71.37} & 1.99& \textbf{59.36} &
\textbf{90.02}& 0.5 & \textbf{50.69} & 
\textbf{90.33}& 0.2& \textbf{59.98} \\
\hline
\end{tabular}
%\vspace{-0.3cm}
\end{table*}

\begin{table*}
\centering
\caption{Continual semantic segmentation results on Mapillary Vistas~2.0 in mIoU (\%). Tasks defined as $\mathcal{C}^1$-$\mathcal{C}^T$($T$ tasks) and $h$ class hierarchy increments. The relative decrease in performance on $\mathcal{C}^1$ is recorded in (\%).}
\label{tab:mapap}
\setlength{\tabcolsep}{3pt}
\begin{tabular} {l|lcc|lcc|lcc|lcc}
% {l|p{0.7cm}p{1.2cm}p{0.7cm}|p{0.7cm}p{1.2cm}p{0.7cm}|p{0.7cm}p{1.2cm}p{0.7cm}|p{0.7cm}p{1.2cm}p{0.7cm}}
 \toprule
 & \multicolumn{3}{c|}{\textbf{51-30 (3 tasks)}} & \multicolumn{3}{c|}{\textbf{71-10 (5 tasks)}} & \multicolumn{3}{c|}{\textbf{39-84 (2 tasks)h}} & \multicolumn{3}{c}{\textbf{39-21 (5 tasks)h}} \\
 \cmidrule{2-13}
\textbf{Method} & 1-51 & $\downarrow$1-51$_{0}$ [\%] &  all & 1-71 & $\downarrow$1-71$_{0}$ [\%] & all & 1-39 & $\downarrow$1-39$_{0}$ [\%] & all & 1-39 & $\downarrow$1-39$_{0}$ [\%] & all \\
\midrule
PLOP~\cite{Douillard2020PLOPLW} & 20.83 & \textbf{17.96} & 14.59 & 18.12 & 27.17 & 13.83 & 19.15 & 44.38& 9.51 & 17.79 &77.37 & 6.64 \\
MiB~\cite{cermelli2020mib} & 16.72 & 34.15 & 13.77 & 15.10 & 39.31 & 12.58 & 19.38 & 43.71 & 11.64 & 16.49 & 52.11 & 8.86 \\
MiB + AWT~\cite{goswami2023attribution} & 18.33 & 27.81 & 15.89 & 15.78 & 36.58 & 13.91 & 19.76 & 42.61 & 15.47 & 17.75 & 48.45 & 12.84 \\
DKD~\cite{baek2022_dkd} & \textbf{25.49} & 22.48 & 18.74 & \textbf{22.71} & 19.27 & 18.65 & \textbf{24.24} & 36.69 & 9.15 & \textbf{28.04} & 26.77 & 8.06 \\
MicroSeg~\cite{zhang22_microseg} & 9.39 & 28.54 & 6.62 & 8.38 & 36.71 & 6.65 & 12.70 & 42.92 & 4.41 & 13.69 & 38.47 & 4.08 \\ 
\midrule
\net~(Ours) & 23.76 & 18.88 & \textbf{20.35} &
22.10 & \textbf{16.54} & \textbf{19.20} 
& 24.16 & \textbf{32.94} & \textbf{21.94}
& 26.67 & \textbf{25.98} & \textbf{23.35} \\
\hline
\end{tabular}
\vspace{-0.2cm}
\end{table*}

\begin{table*}[t]
\centering
\caption{Influence of different class taxonomy trees ($\mathcal{H}$) for performance on incremental classes over training increments on Cityscapes in mIoU (\%).}
\label{tab:tree_sup2}
\begin{tabular}
{l|cccccc|cccccc}
 \toprule
 & \multicolumn{12}{c}{\textbf{14-1 (6 tasks)}} \\
  & \multicolumn{6}{c}{$\mathcal{H}_4$} & \multicolumn{6}{c}{$\mathcal{H}_6$}\\
\cmidrule{2-13}
\textbf{Task} & 15 & 16 & 17 & 18 & 19 & all & 15 & 16 & 17 & 18 & 19 & all \\
\midrule
2 & 48.70 & - & - & - & - & 67.93 &
61.89 & - & - & - & - & 69.10 \\
3 & 34.61 & 51.43 & - & - & - & 65.83 &
58.69 & 66.11 & - & - & - & 68.16 \\
4 & 15.04 & 8.25 & 14.46 & - & - & 59.25 & 
31.67 & 31.48 & 18.12 & - & - & 61.84 \\
5 & 6.26 & 3.47 & 14.39 & 59.42 & - & 58.05 &
27.31 & 31.17 & 18.17 & 63.13 & - & 61.55 \\
6 & 7.87 & 3.92 & 14.93 & 53.47 & 71.99 & 58.37 & 28.11 & 34.43 & 19.93 & 56.70 & 73.15 & 61.74 \\
\hline
\end{tabular}
\vspace{-0.3cm}
\end{table*}

In this supplementary material, we present extended results on Cityscapes and Mapillary Vistas 2.0 in \secref{sec:quanap}. Further, we show additional illustrations of the applied taxonomic trees in \secref{sec:illu}. \secref{sec:sup_h} and \secref{sec:sup_tt} provide results on additional class taxonomies on Cityscapes and give indications of their associated computational costs. We emphasize the positive impact of our regularization losses with additional experiments on Cityscapes and PascalVOC in \secref{sec:sup_reg}. Last, we show the impact of adding our hierarchical loss function to other baseline methods in \secref{sec:sup_h_b} and further analyze the impact of the margin loss in \secref{sec:sup_m}.

\section{Extended Quantitative Results} \label{sec:quanap}
We perform a detailed analysis of the base IoU scores shown in \secref{sec:quan}. All compared methods achieve different performances after base training which significantly influences the observed performance on base classes after the final incremental step. Thus, we note the relative performance degradation of base classes as a percentage of their initial performance in \tabref{tab:cityap} and \tabref{tab:mapap}. On Cityscapes, we observe that \net~presents the least relative performance drop on incremental settings from the background. For the incremental setting from known classes, DKD~\cite{baek2022_dkd} and MicroSeg\cite{zhang22_microseg} can similarly retain knowledge on the non-incremented base class 'sky' which we reason with their missing plasticity to generalize to novel classes.

We show in \tabref{tab:mapap} that \net~has the lowest relative performance degradation on base classes in three scenarios. Consequently, we reason that DKD~\cite{baek2022_dkd} achieves the highest absolute IoU on base classes for this dataset as a result of its favorable performance during the initial base training. We highlight that lower relative performance degradation represents true knowledge retention capabilities which is the focus of this work. Further, we believe that a stronger model can resolve this problem and highlight that \net~can be readily employed with different backbones.

\section{Illustrations of the applied taxonomic trees} \label{sec:illu}
In this section, we present visualizations of different taxonomic trees for Cityscapes and Mapillary Vista 2.0. In \figref{fig:hh1} and \figref{fig:mh3}, we show the taxonomic trees used in \secref{sec:quan}. Further, we present smaller taxonomic trees in \figref{fig:hh2} and \figref{fig:hh3}. Their respective results are presented in \secref{sec:hh}.

\begin{figure*}[hbt!]
    \centering
    \includegraphics[width=0.9\linewidth]{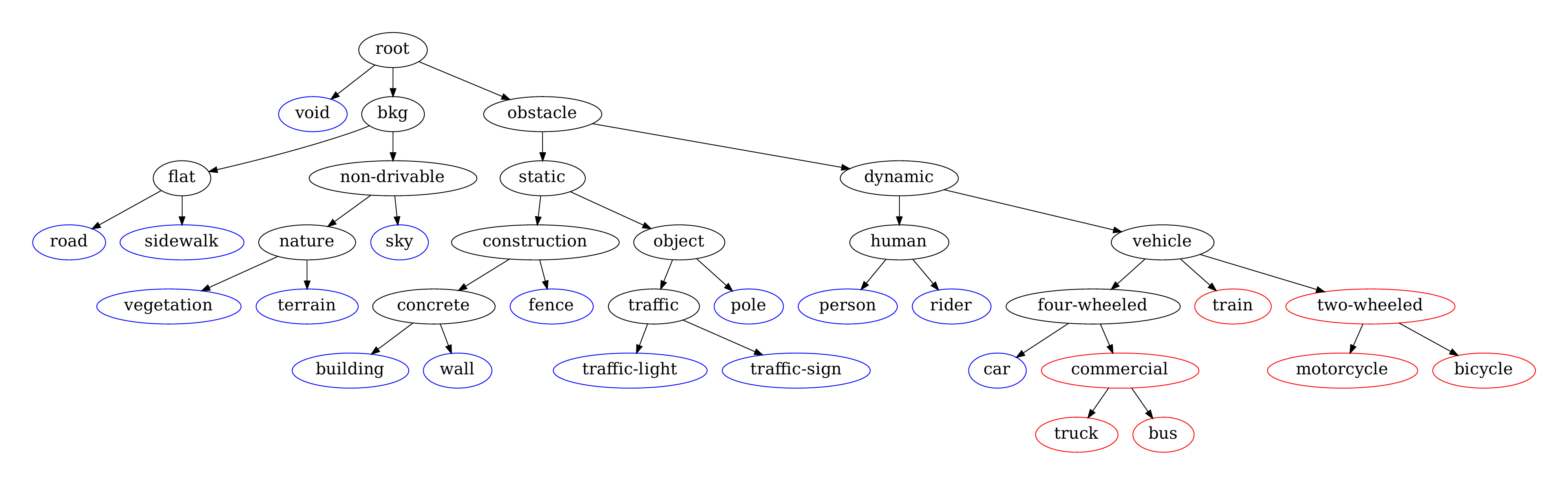}
    \caption{Visualization of the Cityscapes class taxonomic tree  $\mathcal{H}_6$. Base classes ($\mathcal{C}^{1}$) are colored in black (ancestors) and blue (leaf classes) whereas novel classes ($\mathcal{C}^{2:T}$) are colored in red for Cityscapes 14-1 (6 tasks).}
    \label{fig:hh1}
    \vspace{-0.3cm}
\end{figure*}

\begin{figure*}[hbt!]
    \centering
    \includegraphics[width=0.95\linewidth]{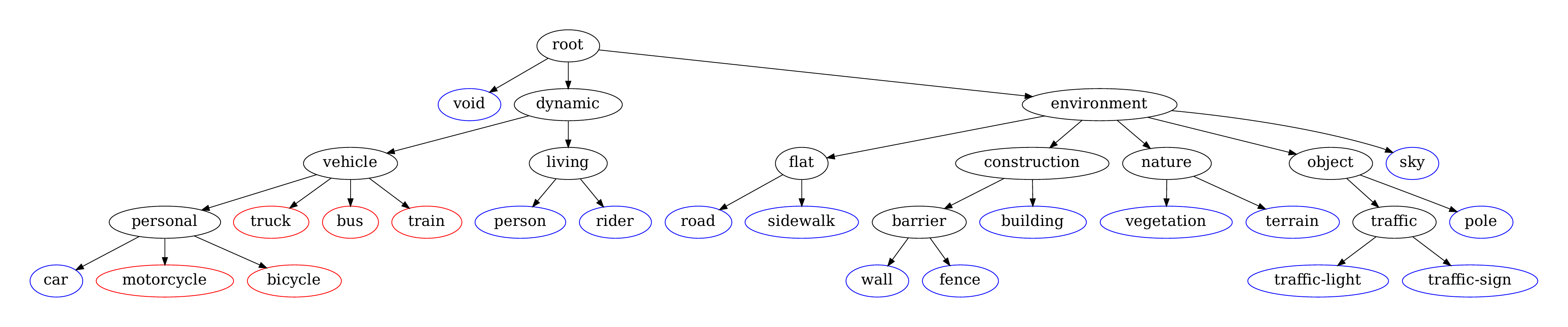}
    \caption{Visualization of the Cityscapes class taxonomic tree $\mathcal{H}_4$. Base classes ($\mathcal{C}^{1}$) are colored in black (ancestors) and blue (leaf classes) whereas novel classes ($\mathcal{C}^{2:T}$) are colored in red for Cityscapes 14-1 (6 tasks).}
    \label{fig:hh2}
    \vspace{-0.3cm}
\end{figure*}

\begin{figure*}[hbt!]
    \centering
    \includegraphics[width=\linewidth]{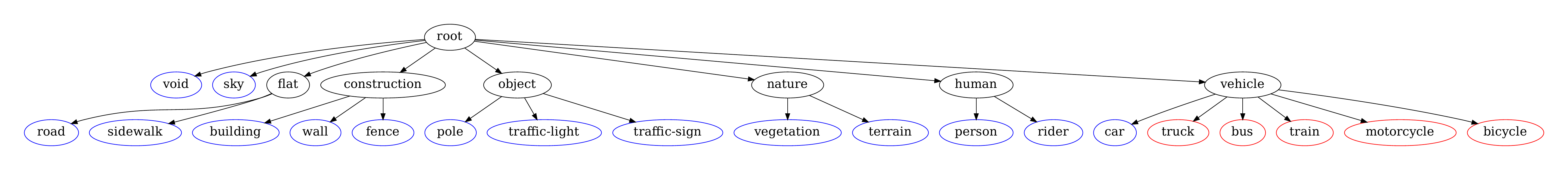}
    \caption{Visualization of the Cityscapes class taxonomic tree $\mathcal{H}_2$. Base classes ($\mathcal{C}^{1}$) are colored in black (ancestors) and blue (leaf classes) whereas novel classes ($\mathcal{C}^{2:T}$) are colored in red for Cityscapes 14-1 (6 tasks).}
    \label{fig:hh3}
    \vspace{-0.3cm}
\end{figure*}

\begin{figure*}[hbt!]
    \centering
  
    \includegraphics[width=1.3\textwidth, angle=90]{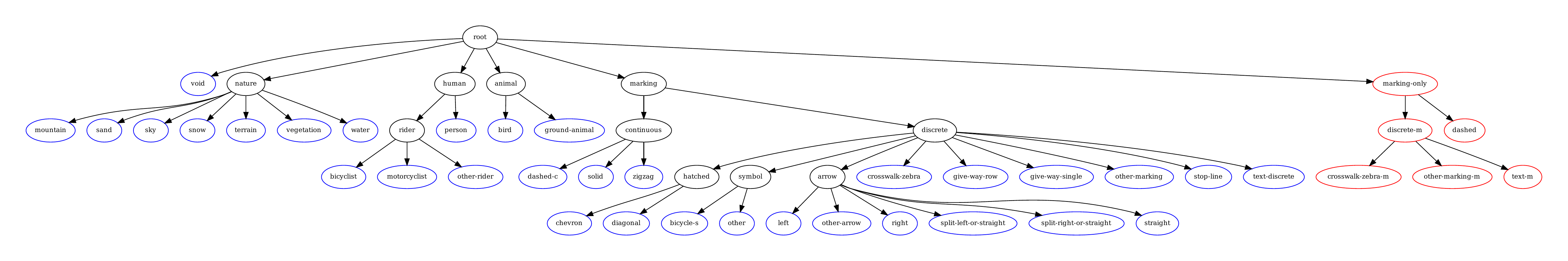}
      \includegraphics[width=1.3\textwidth, angle=90]{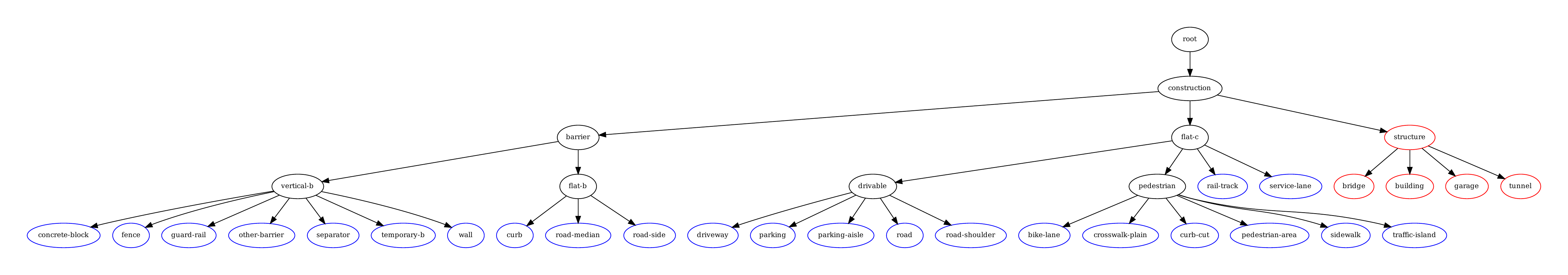}
    \includegraphics[width=1.3\textwidth, angle=90]{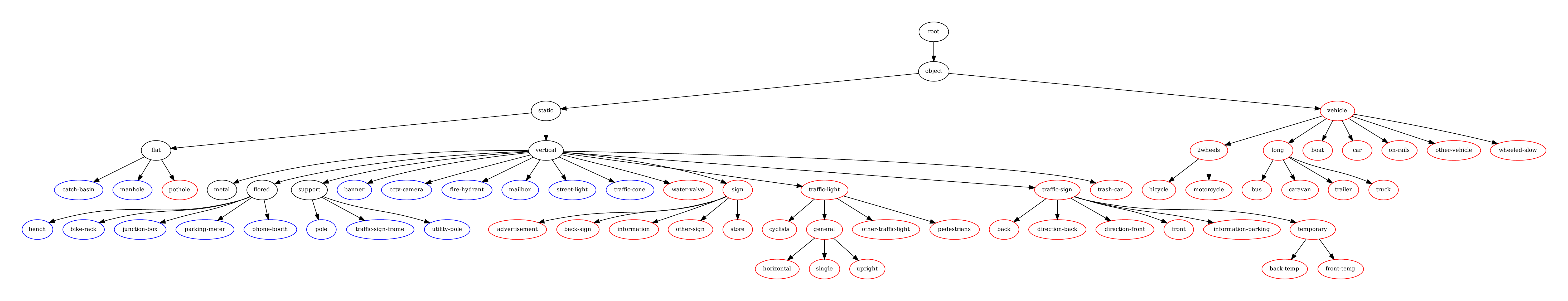}
    \caption{Visualization of the Mappillary Vistas class taxonomic tree. Base classes ($\mathcal{C}^{1}$) are colored in black (ancestors) and blue (leaf classes) whereas novel classes ($\mathcal{C}^{2:T}$) are colored in red for Mapillary 71-10 (5 tasks). The tree is subdivided at the root for visualization purposes.}
    \label{fig:mh3}
    \vspace{-0.3cm}
\end{figure*}

\section{Extended Ablation on Influence of different class taxonomy trees} \label{sec:sup_h}

We complement our findings in the paper that deeper hierarchies prevent forgetting with three further taxonomies shown in \figref{fig:hh3a}, \figref{fig:hh4a} and \figref{fig:hh5a}. For $\mathcal{H}_{3a}$, all predicted classes are at the same hierarchy level (level $3$) in the taxonomy tree. In comparison $\mathcal{H}_{4a}$, extends $\mathcal{H}_3a$ but restricts the hierarchy to have a maximum of $3$ sibling nodes. $\mathcal{H}_{5a}$ allows a maximum of $2$ sibling nodes. 

We show in \tabref{tab:tree_sup} that $\mathcal{H}_{4a}$ outperforms $\mathcal{H}_{3a}$ and $\mathcal{H}_{5a}$. Consequently, we find that having all relevant classes on the same hierarchy level is not required which proves the suitability of our method for incremental learning with unknown taxonomies. Further, we find that class taxonomies that vary the number of siblings between $2-3$, such as $\mathcal{H}_{4a}$ and our best-performing $\mathcal{H}_{6}$ (\figref{fig:hh1}), should be preferred. In summary, we propose that the depth of the hierarchy should be correlated to the number of classes in a way that every internal node has $2-3$ child classes and motivate future research to further explore methods that can predict the best-fitted taxonomic hierarchies.

We further analyze the performance differences between $\mathcal{H}_4$ and $\mathcal{H}_6$ as shown in \tabref{tab:tree}. While the closed-set performance only differs by $0.52$pp, the performance on novel classes is significantly boosted ($12.03$pp) when leveraging $\mathcal{H}_6$. We present the final performance on each individual novel class in \tabref{tab:tree_sup2}. We note that forgetting not only occurs for base classes but also imposes a major challenge for new classes, which are learned at earlier increments. Our proposed hierarchy $\mathcal{H}_6$ significantly aids in achieving a higher performance on novel classes but also preserves this performance. Thus, we conclude that deeper hierarchies prevent forgetting in incremental learning.

\section{Extended Ablation on Computational Costs of different hierarchies} \label{sec:sup_tt}

We measure the computational cost as the total incremental training time on two NVIDIA A40's and average inference time per sample. We present in \tabref{tab:time_rebut} that deeper hierarchies lead to higher training times while the inference time only has a minor increase. This observation can be attributed to the increased complexity of the hierarchical loss with deeper hierarchies.

\section{Extended Ablation on Impact of our Regularization Losses} \label{sec:sup_reg}

 We highlight that pseudo-labeling of the background and our taxonomy-aware loss provide very strong regularization for autonomous driving datasets where certain classes, e.g. sky and road, are commonly observed in every image. 
Consequently, we present results of \net~when only pseudo-labeling the background with a probability of 10\% during incremental training in~\tabref{tab:but_elem_noise}. We disregard the areas of neglected pseudo-labels in all loss function to ensure consistent labeling. We find that the mIoU performance on novel classes drops significantly while their accuracy scores remain high which indicates that novel classes are commonly predicted for ignored image regions. Nevertheless, we find that our regularization losses achieve a performance improvement of $1.37$pp in this experimental setting which can be attributed to an improvement in base and novel classes. Consequently, we highlight that our novel loss functions effectively prevent catastrophic forgetting, which is more pronounced when prior objects appear less frequently in the background.

To strengthen our point, we additionally present results of \net~on the PascalVOC dataset for which prior classes don't commonly appear in other incremental steps in~\tabref{tab:but_elem_voc}. We find that the performance improvement of applying our hierarchical loss amounts $6.49$pp in comparison to a binary cross-entropy on flat labels even if we did not tune the hierarchical loss to achieve a comparable performance after base training. The performance improves by additional $28.78$pp with the application of our two regularization losses. This improvement can be attributed to enhanced scores for both base and novel classes.

\begin{table}
\centering
\caption{Ablation study on the efficacy of various components of \net. All results are reported on Cityscapes in mIoU (\%). All variants apply pseudo-labeling of the background with the old model with a probability of 10\%.}
\label{tab:but_elem_noise}
\begin{tabular}
{p{0.4cm}p{0.4cm}p{0.4cm}|p{0.6cm}p{0.7cm}p{0.5cm}}
 \toprule
 \multicolumn{6}{c}{\textbf{14-1 (6 tasks)}}\\
\textbf{$\mathcal{L}_{hier}$} & \textbf{$\mathcal{L}_{dist}$} & \textbf{$\mathcal{L}_{rel}$} & 1-14 & 15-19 & all \\
\midrule
\checkmark & & & 62.90 & 6.63 & 45.69 \\
\checkmark & & \checkmark & 64.44 & 7.08 & 46.88 \\
\checkmark & \checkmark & & 62.68 & 6.63 & 45.53 \\
\checkmark &\checkmark & \checkmark & \textbf{64.69} & \textbf{7.13} & \textbf{47.06} \\
\hline
\end{tabular}
\end{table}

\begin{table}
\centering
\caption{Ablation study on the efficacy of various components of \net. All results are reported on PascalVOC in mIoU (\%). The setting $1-15_0$ represents the mIoU on base classes at time $t=0$. All variants apply pseudo-labeling of the background with the old model.}
\label{tab:but_elem_voc}
\begin{tabular}
{p{0.4cm}p{0.4cm}p{0.4cm}|p{0.7cm}p{0.6cm}p{0.7cm}p{0.5cm}}
 \toprule
 \multicolumn{7}{c}{\textbf{15-1 (6 tasks)}}\\
\textbf{$\mathcal{L}_{hier}$} & \textbf{$\mathcal{L}_{dist}$} & \textbf{$\mathcal{L}_{rel}$} & 1-15$_{0}$ & 1-15 & 16-20 & all \\
\midrule
& & & 79.78 & 18.21 & 14.15 & 20.30 \\
\checkmark & & & 75.68 & 32.31 & 13.72 & 26.79 \\
\checkmark & & \checkmark & 75.68 & 35.95 & 13.45 & 31.67 \\
\checkmark & \checkmark & & 75.68 & 67.44 & 16.03 & 52.58 \\
\checkmark &\checkmark & \checkmark & \textbf{75.68} & \textbf{68.59} & \textbf{20.66} & \textbf{55.57}\\
\hline
\end{tabular}
\end{table}

\section{Extended Ablation on Impact of Margin Loss}
\label{sec:sup_m}
The margin-based loss in~\cite{liulei24} uses a threshold to determine the minimum distance in its triplet loss function. As suggested by the authors, we tune this parameter according to the depth of the hierarchy. We present further results on the margin loss in \tabref{tab:hr_sub}. While the margin loss shows superior results on closed-set training in Euclidean space (column 1-14$_{0}$), this loss function does not prevent forgetting in incremental learning. Further, we do not record any performance benefits of applying this loss to shallow or deep hierarchies in hyperbolic space. We reason this observation with non-linear increasing distances from the center of the Poincaré ball, whereas one fixed distance threshold is not sufficient to promote further alignment of the hierarchy. This finding provides insights into the construction of hyperbolic loss functions.

\section{Extended Ablation on Impact of our Hierarchical Loss} \label{sec:sup_h_b}
Further, we provide additional experiments in \tabref{tab:other_hier_s} where we add the tree-min loss function to two presented baselines. While the performance of MiB~\cite{cermelli2020mib} is boosted by $1.04$pp, the performance on DKD~\cite{baek2022_dkd} decreased by $5.34$pp. Consequently, we find that naively integrating taxonomies into existing class-incremental learning methods does not automatically result in improved performance. It requires proper design and alignment with knowledge retention methods to make effective use of this supervision.

\begin{table*}[t]
\centering
\caption{Influence of margin loss functions for results on Cityscapes in mIoU (\%). The setting $1-14_0$ represents the mIoU on base classes at time $t=0$.}
\label{tab:hr_sub}
\begin{tabular}
{l|ccc|cccc}
 \toprule
 \multicolumn{8}{c}{\textbf{14-1 (6 tasks)}} \\
Space & $\mathcal{H}$ & Tree-min & Margin Loss & 1-14$_{0}$ & 1-14 & 15-19 & all \\
\midrule
\multirow{2}{*}{\shortstack[l]{Euclidean}} & \multirow{2}{*}{\shortstack[l]{\textbf{$\mathcal{H}_2$}}} & \checkmark & & 68.37 & \textbf{59.37} & 24.93 & \textbf{47.79} \\
& & \checkmark & \checkmark & \textbf{69.69} & 58.85 & \textbf{25.31} & 47.52 \\
\hline
\multirow{4}{*}{\shortstack[l]{Hyperb.}} & \multirow{2}{*}{\shortstack[l]{\textbf{$\mathcal{H}_2$}}} & \checkmark & & 66.46 & 60.86 & 32.57 & 50.75 \\
& & \checkmark & \checkmark & 66.31 & 60.43 & 30.13  & 49.83 \\

& \multirow{2}{*}{\shortstack[l]{\textbf{$\mathcal{H}_6$}}} & \checkmark & & \textbf{74.85} & \textbf{73.03} & \textbf{42.47} & \textbf{61.74}\\
& & \checkmark & \checkmark & 74.20 & 72.97 & 35.31 & 59.91 \\
\hline
\end{tabular}
\end{table*}

\begin{table}
\centering
\caption{Integration of hierarchical supervision to baseline methods on Cityscapes in mIoU (\%).}
\label{tab:other_hier_s}
\begin{tabular}
{l|ccc}
 \toprule
& \multicolumn{3}{c}{\textbf{14-1 (6 tasks)}} \\
\textbf{Method} & 1-14 & 15-19 & all \\
\midrule
DKD~\cite{baek2022_dkd} & 68.83 & 14.70 & 51.86\\ 
DKD$_{hier}$ & 61.01 & 15.27 & 46.52 \\
MiB~\cite{cermelli2020mib} & 66.37 & 14.36 & 50.05\\ 
MiB$_{hier}$ & 68.29 & 13.14 & 51.09\\
\hline
\end{tabular}
\end{table}

\begin{figure*}[hbt!]
    \centering
    \includegraphics[width=\linewidth]{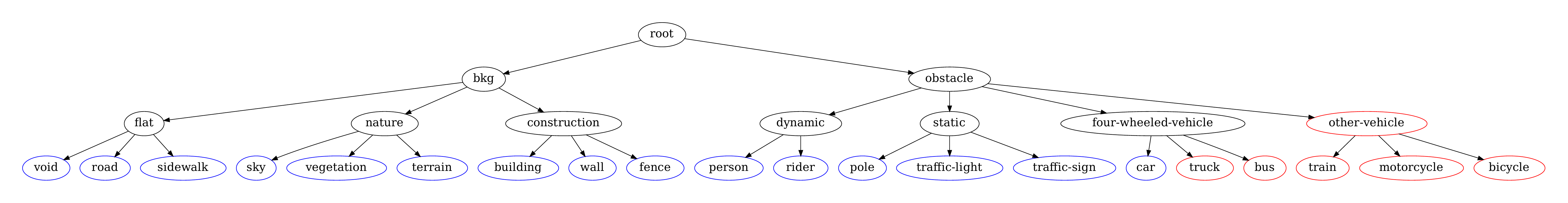}
    \caption{Visualization of the Cityscapes class taxonomic tree $\mathcal{H}_{3a}$. Base classes ($\mathcal{C}^{1}$) are colored in black (ancestors) and blue (leaf classes) whereas novel classes ($\mathcal{C}^{2:T}$) are colored in red for Cityscapes 14-1 (6 tasks).}
    \label{fig:hh3a}
    \vspace{-0.3cm}
\end{figure*}

\begin{figure*}[hbt!]
    \centering
    \includegraphics[width=\linewidth]{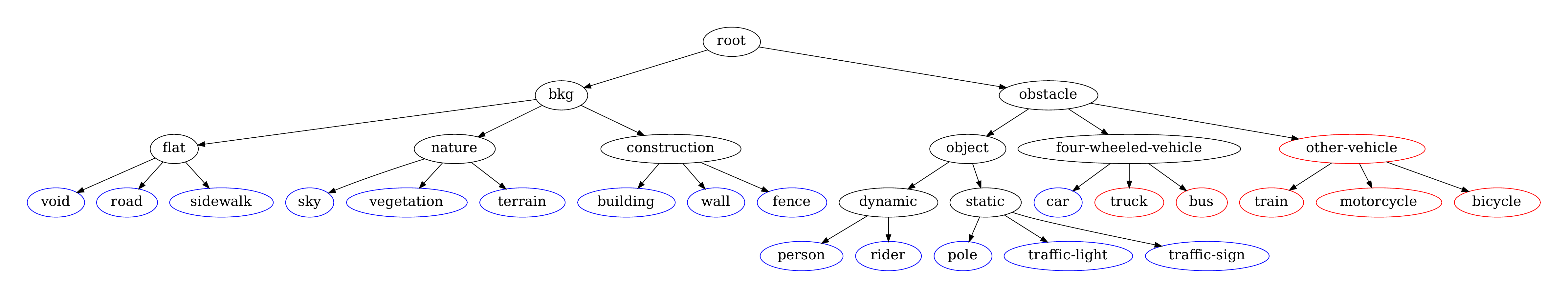}
    \caption{Visualization of the Cityscapes class taxonomic tree $\mathcal{H}_{4a}$. Base classes ($\mathcal{C}^{1}$) are colored in black (ancestors) and blue (leaf classes) whereas novel classes ($\mathcal{C}^{2:T}$) are colored in red for Cityscapes 14-1 (6 tasks).}
    \label{fig:hh4a}
    \vspace{-0.3cm}
\end{figure*}

\begin{figure*}[hbt!]
    \centering
    \includegraphics[width=\linewidth]{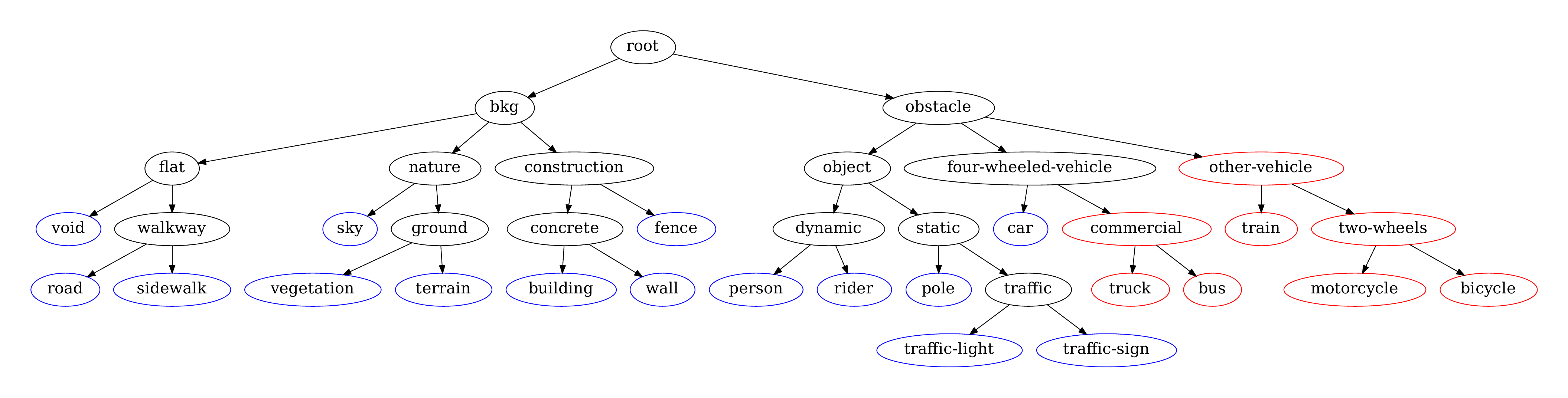}
    \caption{Visualization of the Cityscapes class taxonomic tree $\mathcal{H}_{5a}$. Base classes ($\mathcal{C}^{1}$) are colored in black (ancestors) and blue (leaf classes) whereas novel classes ($\mathcal{C}^{2:T}$) are colored in red for Cityscapes 14-1 (6 tasks).}
    \label{fig:hh5a}
    \vspace{-0.3cm}
\end{figure*}
\newpage
{\footnotesize
\bibliographystyleS{IEEEtran}
% \bibliographyS{references}
}

    \newpage
\fi

% \input{sections/supp.tex}
% \newpage
% {\footnotesize
% \bibliographystyleS{IEEEtran}
% \bibliographyS{references}
% }
\end{document}